\documentclass[lettersize,journal]{IEEEtran}
\usepackage{amsmath,amsfonts,algorithmic,algorithm,array,booktabs,bm,color,cite,epstopdf,graphicx,hhline,mathtools,multirow,pdfpages,soul,textcomp,subcaption,stfloats,url,verbatim,}
\usepackage{graphicx}
\usepackage[table,xcdraw]{xcolor}
\usepackage[colorlinks=true, allcolors=blue]{hyperref}
\DeclarePairedDelimiter\ceil{\lceil}{\rceil}

\begin{document}
\title{SPP-CNN: An Efficient Framework for Network Robustness Prediction}

\author{Chengpei~Wu,~
	Yang~Lou,~
	Lin~Wang,~
	Junli~Li,~
	Xiang~Li,~
	and Guanrong~Chen%
	\thanks{Chengpei~Wu and Junli~Li are with the College of Computer Science, Sichuan Normal University, Chengdu 610066, China, and also with the Key Laboratory of System Control and Information Processing, Ministry of Education, Shanghai 200240, China (e-mail: chengpei.wu@hotmail.com, lijunli@sicnu.edu.cn).}
	\thanks{Yang~Lou is with the Department of Computer Science, National Yang Ming Chiao Tung University, Hsinchu 300, Taiwan (e-mail: felix.lou@ieee.org).}
	\thanks{Lin Wang is with the Department of Automation, Shanghai Jiao Tong University, Shanghai 200240, China, and also with the Key Laboratory of System Control and Information Processing, Ministry of Education, Shanghai 200240, China (e-mail: wanglin@sjtu.edu.cn).}%
	\thanks{Xiang Li is with the Institute of Complex Networks and Intelligent Systems, Shanghai Research Institute for Intelligent Autonomous Systems, Tongji University, Shanghai 201210, and also with the Department of Control Science and Engineering, Tongji University, Shanghai 200240, China (e-mail: lix2021@tongji.edu.cn).}%
	\thanks{Guanrong Chen is with the Department of Electrical Engineering, City University of Hong Kong, Hong Kong, China (e-mail: eegchen@cityu.edu.hk).}%
	\thanks{This manuscript has been submitted for potential publication.}
	\thanks{(\textit{Chengpei Wu and Yang Lou contributed equally to this work.})}
	\thanks{(\textit{Corresponding author: Yang Lou})}
	}
	\markboth{Journal of \LaTeX\ Class Files,~Vol.~XX, No.~YY, May~2023}%
	{Lou\MakeLowercase{\textit{et al.}}: A Sample Article Using IEEEtran.cls for IEEE Journals}
	\maketitle
	
\begin{abstract}
This paper addresses the robustness of a network to sustain its connectivity and controllability against malicious attacks. This kind of network robustness is typically measured by the time-consuming attack simulation, which returns a sequence of values that record the remaining connectivity and controllability after a sequence of node- or edge-removal attacks. For improvement, this paper develops an efficient framework for network robustness prediction, the spatial pyramid pooling convolutional neural network (SPP-CNN). The new framework installs a spatial pyramid pooling layer between the convolutional and fully-connected layers, overcoming the common mismatch issue in the CNN-based prediction approaches and extending its generalizability. Extensive experiments are carried out by comparing SPP-CNN with three state-of-the-art robustness predictors, namely a CNN-based and two graph neural networks-based frameworks.	Synthetic and real-world networks, both directed and undirected, are investigated. Experimental results demonstrate that the proposed SPP-CNN achieves better prediction performances and better generalizability to unknown datasets, with significantly lower time-consumption, than its counterparts.
\end{abstract}

\begin{IEEEkeywords}
Complex network, robustness, convolutional neural network, spatial pyramid pooling, prediction.
\end{IEEEkeywords}

\section{Introduction}

\IEEEPARstart {T}{he} study of complex networks enhances our understanding of various real-world systems, ranging from natural to engineering, to biological, and to social networks\cite{Barabasi2016NS,Newman2010N,Chen2014Book}. Scientific investigations intend to reveal the essence and characteristics of networks, while engineering studies focus on their functions and applications, regarding such as connectivity\cite{Schneider2011PNAS,Ellens2013arXiv,Freitas2022TKDE}, controllability\cite{Liu2011N,Yuan2013NC,Posfai2013SR,Menichetti2014PRL,Wang2016AUTO,Wang2017RSPTA,Xiang2019CSM,Wu2020TCASI}, data transmission and communication abilities\cite{Lu2019CL,Mburano2021ISNCC}, and so on. Today, random failures and malicious attacks take place in many engineering and technological applications, which significantly degrade or even destroy the network normal functions. Therefore, it has become important and necessary to strengthen such network functions against  attacks and failures\cite{Holme2002PRE,Shargel2003PRL,Cohen2010Book,Schneider2011PNAS,Liu2012PO,Bashan2013NP,Ellens2013arXiv,Chan2016DMKD,Liu2017FCS,Freitas2022TKDE}. This leads to the concern of \textit{network robustness}, which has different meanings in different scenarios, and here it refers to the ability of a network to sustain its normal functions when a fraction of the network nodes and/or edges failure due to attacks. In this paper, network robustness is studied with respect to both connectivity and controllability against destructive attacks and failures in the forms of node-removals.

The enhancement of network robustness depends on some reliable and efficient measures of the defined robustness\cite{Ellens2013arXiv,Freitas2022TKDE}. In general, network robustness can be measured in both \textit{a priori} and \textit{a posteriori} manners. \textit{A priori} measures are referred to specific network features that can be evaluated and calculated without performing attack simulations. Widely-used \textit{a priori} robustness measures include topological measures, for example betweenness centrality\cite{Freeman1977Soc} and clustering coefficient\cite{Watts1998N}, and spectral measures include such as natural connectivity\cite{Wu2010CPL}, algebraic connectivity\cite{Fiedler1973CMJ}, effective resistance\cite{Klein1993JMC}, and so on. Network spectra can be calculated using the network adjacency matrix and Laplacian matrix\cite{Gavili2017TSP}. Although \textit{a priori} measures are easy-to-access with lower computational cost and complexity\cite{Freitas2022TKDE,Chan2016DMKD}, they have limited scopes of applications\cite{Yamashita2019COMPSAC}.

In contrast, \textit{a posteriori} robustness measures have intuitively clear meanings with a wider range of applicability, namely, \textit{a posteriori} measures are applicable to any type of networks under any kind of attacks, and they return distinguishable measure values with respect to different attack strategies. Therefore, the practical \textit{a posteriori} measures remain as the main approach in robustness studies for many real-world applications. \textit{A posteriori} measures are quantified by recording a sequence of values that represent the remaining functionality (here, connectivity and controllability) of the network after a series of node- or edge-removal attacks. Evaluation of \textit{a posteriori} measures is generally time-consuming, however, not only due to the iterative node- or edge-removal processes, but also because of the recalculation of the concerned network functions, such as the size of largest connected component\cite{Schneider2011PNAS}, the number of needed driver nodes\cite{Liu2011N,Yuan2013NC}, and the number of communicable node pairs\cite{Lu2019CL,Mburano2021ISNCC}.

In practice, it is not always necessary to calculate the exact values of the network robustness as done in attack simulation. For example, during an optimization process, network robustness is reevaluated after each structural disturbance, e.g., edge rewiring\cite{Zeng2012PRE,Schneider2013SR,Lou2021TCASII}. It is very time-consuming to perform an attack simulation each time, especially for large-scale networks. Therefore, robustness prediction approaches are more desirable in such cases, which significantly reduce the cost and time complexity\cite{Wang2021TEVC}.

Network robustness prediction can be achieved by using either analytical or computational methods. In so doing, the time complexity is either constant\cite{Lou2022IJCAS} or increasing but significantly slower than that of the attack simulation\cite{Lou2022TCYB}. Analytical approximations have a much narrower scope of applications to, e.g., the controllability robustness under random edge-attacks\cite{Sun2019ICSRS,Sun2021TNSM,Lou2022IJCAS}. In contrast, machine learning algorithms, such as neural networks and random forest schemes, have no such limitation\cite{Dhiman2021MLN}. By combining several machine learning algorithms together, a surrogate ensemble can be formed for robustness prediction through an optimization process\cite{Wang2021TEVC}.

Regarding neural networks-based frameworks, deep neural networks are more powerful than canonical machine learning algorithms for efficiently processing network data. Successful application examples include critical node identification using deep reinforcement learning\cite{Fan2020NMI} and graph attention networks\cite{Grassia2021NC}. The convolutional neural network (CNN)\cite{Schmidhuber2015NN}-based prediction processes network data as gray-scale images\cite{Lou2022TCYB}, thereby fast approximating the robustness performances against different attacks. Prior knowledge is useful to further improve the prediction performances\cite{Lou2021TNNLS}. This straightforward approach requires downsampling and upsampling for the input data that are smaller and larger than the fixed input size of CNN, however, thus information distortion may be severe if the input network size is very large or very small. Graph neural networks (GNN)\cite{Kipf2016arXiv,Hamilton2020Book} are able to compress and unify higher-dimensional network raw data to lower-dimensional representations. Therefore, GNN-based approaches are more tolerable to the input data-size changes, and so have better prediction performances\cite{Lou2022arXiv}. Taking the connectivity robustness prediction as an example, the CNN-based approach needs an average run time that is $3.48\%$ of the attack simulation, while the GNN-based approach requires up to $82.8\%$ of the attack simulation\cite{Lou2022arXiv}.

In this paper, to overcome the aforemention mismatch issue between the various input sizes and the fixed input size in the CNN-based processing, and also to achieve a balance between the flexibility of processing different input sizes and the approximation speed in computation, a spatial pyramid pooling (SPP)\cite{He2015TPAMI} layer is installed into the network robustness predictor, which is demonstrated to be superior to the other CNN-based and GNN-based frameworks.

Specifically, the work and contributions of this paper are summarized as follows:
\begin{itemize}
	\item[1)] SPP-CNN is proposed, which has a wider tolerance to different input-data sizes than the CNN- and GNN-based approaches\cite{Lou2022arXiv}, while maintaining fast approximation speed like the CNN-based approaches\cite{Lou2022TCYB}.
	\item[2)] SPP-CNN shows stronger generalizability than the other approaches on predicting the network robustness for datasets with unseen topologies and sizes.
	\item[3)] SPP-CNN demonstrates better performances than the other approaches on predicting the robustness of real-world networks, with consistent advantages for both synthetic and real-world networks.
\end{itemize}

The rest of the paper is organized as follows.
Section \ref{sec:pre} reviews several measures of the network robustness against destructive node-removal attacks, and both CNN-based and GNN-based prediction frameworks.
Section \ref{sec:spp} introduces the proposed SPP-CNN.
Section \ref{sec:exp} presents extensive empirical experiment results with comparison and analysis.
Section \ref{sec:end} concludes the investigation.

\section{Preliminaries} \label{sec:pre}

In this paper, network robustness is considered from two specific aspects, namely the \textit{a priori} connectivity robustness and controllability robustness, while other \textit{a posteriori} robustness measures such as communication robustness can be investigated in a similar manner. Only node-removal attacks is discussed, while edge-attacks can also be studied in the same way. Three state-of-the-art network robustness predictors are reviewed, namely the CNN-based robustness predictor (CNN-RP)\cite{Lou2022TCYB}, PATCHY-SAN\cite{Niepert2016ICML}, and the learning feature representation-based predictor (LFR-CNN)\cite{Lou2022arXiv}. Both PATCHY-SAN and LFR-CNN consist of a GNN module for graph representation learning, followed by a CNN for robustness performance prediction (regression).

\subsection{Robustness Measures}\label{subsub:nec}

\subsubsection{Connectivity Robustness}

In this paper, connectivity robustness refers to the ability of a network to maintain its connectivity against destructive node-removal attacks, which is widely measured by counting the size of the largest connected component (LCC) in the network after an attack. If an undirected network is connected or a directed network is weakly connected, then the LCC is the network itself; otherwise, a connected component that includes the largest number of nodes is an LCC. Connectivity robustness is evaluated by the following index:
\begin{equation}\label{eq:nlc}
	R_1=\sum_{i=0}^{N-1}{r_{1}(i)}=\sum_{i=0}^{N-1}{\frac{N_L(i)}{N-i}}\,,
\end{equation}
where $N$ is the number of nodes in the given network before being attacked; $i$ is the total number of nodes that have been removed from the network; $N_L(i)$ is the number of nodes in the remaining LCC; therefore, $r_{1}(i)$ is the density of nodes remaining in the $i$th LCC. The denominator $N-i$ ensures $R_1\in(0,1]$, but if the denominator is replaced by $N$ then $R_1\in(0,0.5]$ as used in \cite{Schneider2011PNAS}.

The series of values $r_{1}(i)$ can be plotted to show a \textit{connectivity curve}, for which the scalar $R_1$ reflects the overall connectivity robustness against node-removal attacks: a higher $R_1$ value indicates a better connectivity robustness.

\subsubsection{Controllability Robustness}\label{subsub:con}

Controllability robustness reflects the ability of a networked system to maintain or regain its controllability with the lowest control cost, e.g., the minimum number of driver nodes (DN) that are needed to add to the network after the attack. For a general linear time-invariant (LTI) networked system $\dot{{\bf x}}=A{\bf x}+B{\bf u}$, where ${\bf{x}}\in\mathbb{R}^N$ represents the state vector; ${\bf{u}}\in\mathbb{R}^b$ represents the control input; $A\in\mathbb{R}^{N\times N}$ and $B\in\mathbb{R}^{N\times b}$ are constant matrices. This LTI system is \textit{state controllable} if and only if there exist a control input $\bf{u}$ that can drive the state $\bf{x}$ to move from any initial state to any target state in the state space in finite time. The state controllability can be determined by checking whether the controllability matrix $[B\ AB\ A^2B\ \cdots A^{N-1}B]$ has a full row-rank\cite{Chen1998Book}. The concept of \textit{structural controllability} is a slight generalization of the state controllability, to deal with two parameterized matrices $A$ and $B$, in which the parameters characterize the structure of the underlying system in the sense that if there are specific parameter values that can ensure the system to be state controllable then the system is said to be structurally controllable.

For a network with many LTI systems, any node system with control input is a ND. The minimum number of NDs needed to retain the (state or structural) controllability of the network can be determined by using either the minimum inputs theorem (MIT)\cite{Liu2011N} for directed networks or the exact controllability theorem (ECT)\cite{Yuan2013NC} for both directed and undirected networks, which are defined as follows:
\begin{equation}\label{eq:2nd}
	N_D=
	\begin{cases}
		\text{max}\{1, N-|E|\},& \text{using MIT\cite{Liu2011N}},\\
		\text{max}\{1, N-\text{rank}(A)\},& \text{using ECT\cite{Yuan2013NC}},\\
	\end{cases}
\end{equation}
where $|E|$ represents the number of edges in the maximum matching $E$, which is a basic concept in graph theory\cite{Liu2011N}. Under a sequence of node-removal attacks, the controllability robustness is measured by
\begin{equation}\label{eq:nd}
	R_2=\sum_{i=0}^{N-1}{r_2(i)}=\sum_{i=0}^{N-1}{\frac{N_D(i)}{N-i}}\,,
\end{equation}
where $N_D(i)$ is the number of DNs and $r_2(i)$ is the density of DNs, which are needed to retain the network controllability after a total of $i$ nodes have been removed by the attack. Similarly, the series of values $r_2(i)$ can be plotted to show a \textit{controllability curve}, where $R_2$ measures the overall controllability robustness: a lower $R_2$ values indicates a better connectivity robustness.

\subsection{Network Robustness Predictors}

There are three commonly-used network robustness predictors, namely CNN-RP\cite{Lou2022TCYB}, PATCHY-SAN\cite{Niepert2016ICML}, and FR-CNN\cite{Lou2022arXiv}. Basic principles as well as the pros and cons of these approaches are reviewed and discussed in this subsection.

\begin{figure}[htbp]
	\centering \includegraphics[width=\linewidth]{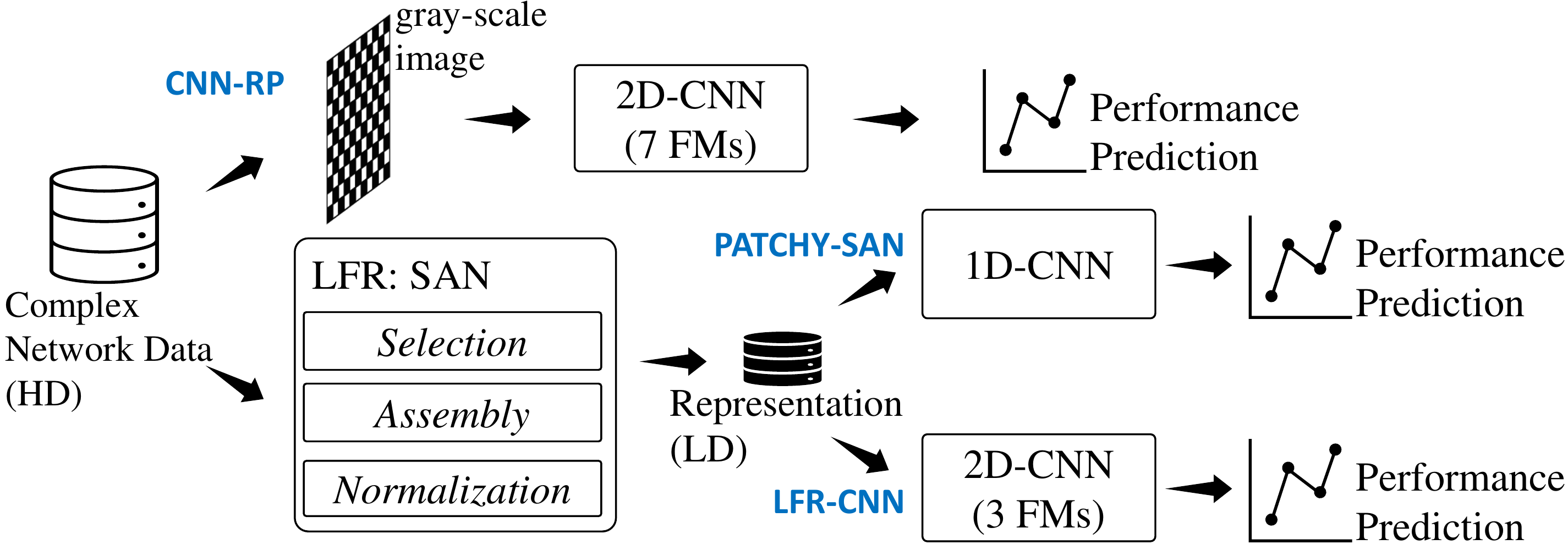}
	\caption{General frameworks of CNN-RP, PATCHY-SAN, and LFR-CNN. CNN-RP processes network data as gray-scaled images, while for PATCHY-SAN and LFR-CNN, the LFR module performs the selection, assembly, and normalization (SAN) operations, which compress higher-dimensional (HD) network data to be lower-dimensional (LD) representations.}\label{fig:gen_flow}
\end{figure}

The general structures of CNN-RP, PATCHY-SAN, and LFR-CNN are visualized in Fig. \ref{fig:gen_flow}. The input is the adjacency matrix, which could also be a Laplacian matrix or other representations, and the output is the predicted robustness performance.

\subsubsection{CNN-RP}

\begin{figure}[htbp]
	\centering \includegraphics[width=\linewidth]{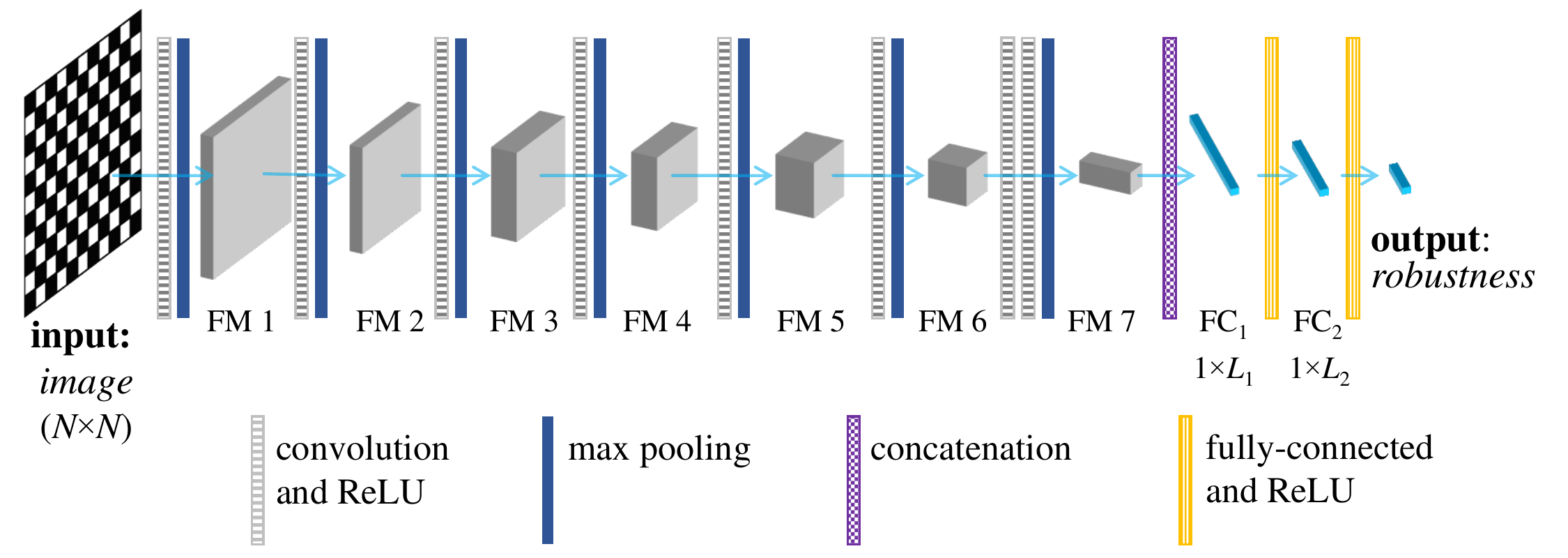}
	\caption{CNN structure in CNN-RP. The input is an adjacency matrix; the output is an $N$-vector. Seven feature map (FM) groups are generated with $N_i=\ceil{N/2^{(i+1)}}$, for $i=1,2,\ldots,7$. Concatenation layer reshapes the data to be a vector, from FM 7 to FC$_1$. FC$_1$=$512N_7^{2}$ and FC$_2$=$4096$\cite{Lou2022TCYB}. }\label{fig:cnn}
\end{figure}

The structure of CNN-RP\cite{Lou2022TCYB,Lou2021TNSE}, using a VGG-structured CNN\cite{Simonyan2014arXiv}, is shown in Fig. \ref{fig:cnn}. Adjacency matrices are treated as gray-scale images and processed by CNN directly. Classification and regression tasks are completed using such an image-processing mechanism\cite{Lou2021TNNLS}.

The output of CNN-RP is an $N$-vector, which represents a connectivity or controllability curve, denoted by $\hat{v}$. The mean-squared error between the predicted curve $\hat{v}$ and the true curve $v$ is used as the loss function:
\begin{equation}\label{eq:lf}
	\mathcal{L} = \frac{1}{N} \sum_{i=1}^{N}||\hat{v}(i)-v(i)||\,,
\end{equation}
where $\hat{v}(i)$ and $v(i)$ represent the predicted and the true connectivity or controllability values, respectively; $i$ is the total number of nodes that have been removed from the network; $||\cdot||$ is the Euclidean norm. The true values $v(i)$ are obtained by performing attack simulation. The training process aims at adjusting the internal parameters, aiming at minimizing $\mathcal{L}$.

\subsubsection{PATCHY-SAN and LFR-CNN}\label{subsub:lfr}

As shown in Fig. \ref{fig:gen_flow}, PATCHY-SAN and LFR-CNN share the same LFR module, which converts higher-dimensional network data to lower-dimensional representations. This conversion is achieved by using a series of three operations: selection, assembly, and normalization (SAN).

First, the $N$ network nodes are sorted according to their importance, which can be quantified by certain node centrality measure such as node degree or node betweenness. A total of $W$ most important nodes are first selected. For each selected node, a receptive field of size $g$ is created based on its neighboring information. If $N<W$, then all-zero receptive fields are added for padding. Next, a breadth-first search is conducted to construct the neighborhood field for each selected node. Finally, a normalization step converts and normalizes the neighboring field for each selected node to an embedded vector with a uniform length $gh$, where $h$ is the number of attributes used for the neighboring nodes. The first element in the resultant normalized vector represents the root node, followed by the neighboring nodes sorted according to their centrality measures.

To this end, an $N$-node network represented by an $N^2$ adjacency matrix has been represented by a $W\times (gh)$ matrix, where there are $W$ receptive fields and each receptive field is presented by a $1\times gh$ vector. Since $g\ll N$ and $h\ll N$, this procedure generates lower-dimensional learned feature representations from higher-dimensional raw network data. Then, the lower-dimensional representations are processed by CNN to predict the robustness performances of the given networks.

For PATCHY-SAN, a shallow 1D-CNN structure is employed, while for LFR-CNN a VGG-structured\cite{Simonyan2014arXiv} 2D-CNN with 3 feature maps (FMs) is used. 
The numbers of internal parameters to be adjusted during training are $5.1\times10^5$ and $6.0\times10^6$ for PATCHY-SAN and LFR-CNN, respectively.

The same mean-squared error between the predicted curve $\hat{v}$ and the true curve $v$ as shown in Eq. (\ref{eq:lf}) is used as the loss function for both PATCHY-SAN and LFR-CNN.

\subsection{Error Measures}

Let ${\bf{v_t}}=\{v_t(i)\}_{i=0}^{N-1}$ and ${\bf{v_p}}=\{v_p(i)\}_{i=0}^{N-1}$ be the true and the predicted robustness curves obtained by attack simulation, respectively. The \textit{prediction error} $\xi$ is calculated by
\begin{equation}\label{eq:avg_xi}
	\xi=\frac{1}{N}\sum_{i=0}^{N-1}{\xi(i)}\,,
\end{equation}
where $\xi(i)=|v_t(i)-v_p(i)|$, $i=0,1,...,N-1$. Given two prediction errors obtained by two different robustness predictors, denoted by ${\xi}_1$ and ${\xi}_2$, if ${\xi}_1<{\xi}_2$ then the first predictor performs better than the second.

\section{Spatial Pyramid Pooling}\label{sec:spp}

A CNN consists of convolutional layers and fully-connected layers, as shown in Fig. \ref{fig:cnn}. The convolutional layers, which work in a sliding-window manner, are flexible with different input sizes, while the nature of the fully-connected layers requires a predefined fixed size of input. Therefore, CNNs require a fixed size for the input data, e.g., $224 \times 224$. For image processing tasks such as classification and object detection\cite{Simonyan2014arXiv,Szegedy2015CVPR,He2016CVPR},
cropping\cite{Zeiler2014ECCV} and warping\cite{Girshick2014CVPR}, the images may be fit to the required fixed input size. When complex network data are used as the input to CNNs, upsampling or subsampling is suitable\cite{Lou2022TCYB}, since cropping and warping may discard or distort some specific marginal regions of an image, while upsampling and subsampling cause uniformly random information loss or distortion. However, resizing of network data not only will change the true network size, but also will change the true topology, by removing existing nodes or adding dummy nodes. But the resized network data can be misleading, for example, when hub nodes are deleted.

On the other hand, the GNN-based approaches can effectively reduce the information loss by extracting neighboring information from the important nodes, as discussed in Subsection \ref{subsub:lfr}. The representation learning process converts the input higher-dimensional data of any size to a regulated fixed-sized lower-dimensional representation, and then passes it to a CNN. This process significantly widen the application range of the robustness predictor to different network sizes\cite{Lou2022arXiv}. However, the GNN-based approaches are significantly slower than the CNN-based approaches, where the representation learning process is the most time-consuming part.

Embedding a spatial pyramid pooling (SPP)\cite{He2015TPAMI} layer into CNN, in between the convolutional layers and fully-connected layers, brings some benefits: it brings a buffer layer from the flexible convolutional layers to the fixed fully-connected layers. SPP is developed from the canonical spatial pyramid matching algorithm in computer vision\cite{Grauman2005ICCV,Lazebnik2006CVPR}, which statistically counts the feature distributions of images from multiple scales, such that better recognition performances can be achieved. The spatial pyramid matching algorithm sets a number of spatial bins, and then local features within each individual spatial bin are captured statistically. The representation of the whole image consists of local features from different spatial bins.

As an extension of the spatial pyramid matching algorithm, SPP uses max pooling instead of statistical counting to local feature learning and resizing. Given input images of different sizes, the convolution layers are able to process all the information and then generate feature maps of different sizes. The SPP layer then transforms these feature maps to fixed-length representations.

\begin{figure*}[htbp]
	\centering \includegraphics[width=1\linewidth]{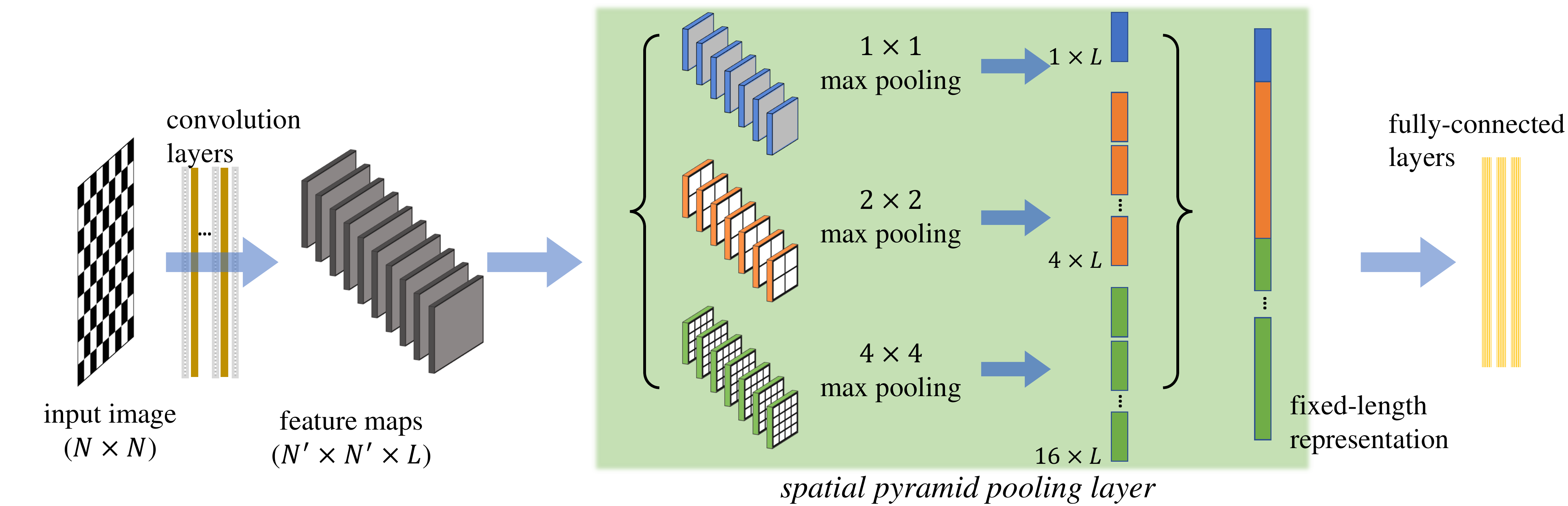}
	\caption{The convolutional neural network structure with a spatial pyramid pooling layer. The detailed structure of convolutional layers are shown in Fig. \ref{fig:spp_cnn}. } \label{fig:spp}
\end{figure*}

As shown in Fig. \ref{fig:spp}, an $N\times N$ input image results in $L$ $N'\times N'$ feature maps, where $L$ is the number of filters in the last convolutional layer.
In the SPP layer, feature maps are divided into 3 different levels of spatial bins, of sizes $1\times1$, $2\times2$, and $4\times4$, which are processed by max pooling with corresponding sizes. Then, a representation vector of size $pL$ is generated as the output of the SPP layer. Here, both $L$ and $p$ are pre-defined hyperparameters. As a result, for the input image of any size, a fixed length $pL$-vector is generated as the input to the fully-connected layers. In this paper, three pyramid pooling levels are used, with sizes of $1\times1$, $2\times2$ and $4\times4$, respectively. It has been empirically verified that the performance of the SPP layer is insensitive to different settings of the pyramid bins\cite{He2015TPAMI}.

\begin{figure}[htbp]
	\centering \includegraphics[width=\linewidth]{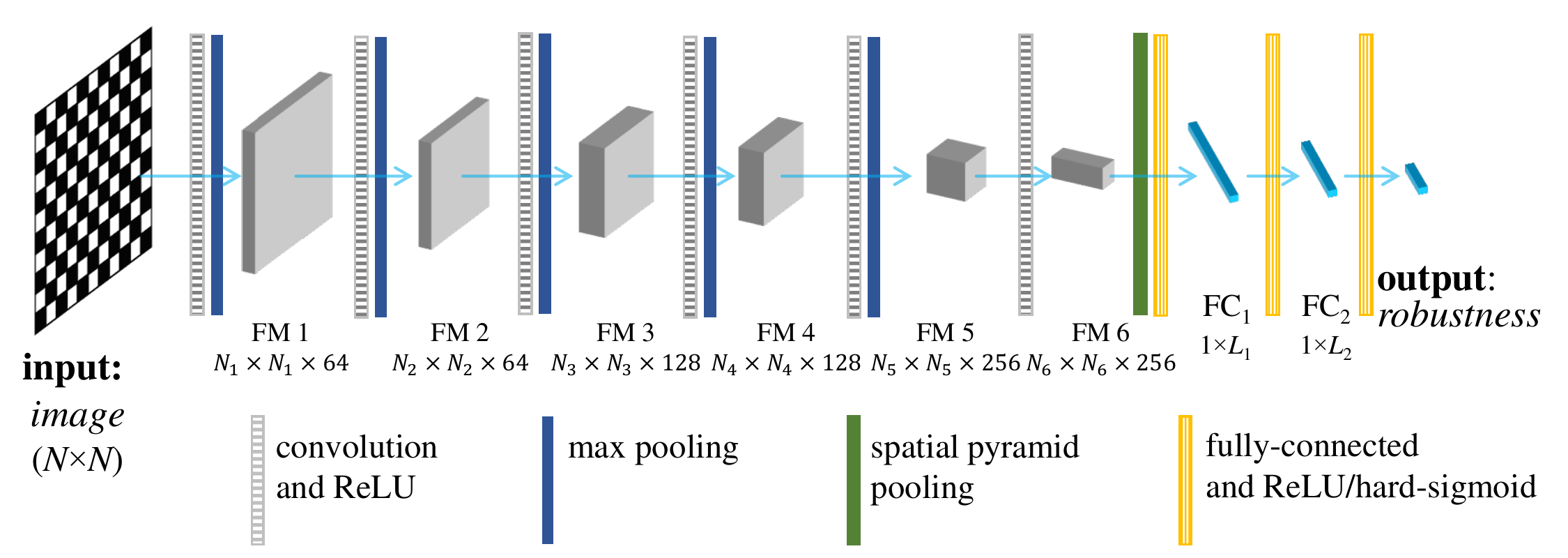}
	\caption{CNN structure of SPP-CNN. The spatial pyramid pooling layer is installed between the convolutional layers and fully-connected layers. Hard-sigmoid is installed in the last fully-connected layer, while in other layers ReLU is installed.}\label{fig:spp_cnn}
\end{figure}

By installing such an SPP layer in CNN, an SPP-CNN is constructed for network robustness prediction. Fig. \ref{fig:spp_cnn} shows the structure of SPP-CNN, which consists of 6 convolutional layers, 1 SPP layer, and 3 fully connected layers. This structure is able to predict the robustness performances of networks with hundreds to thousands of nodes. Specifically, $p=21$ represents the total number of bins, and $L=256$ denotes the number of filters in the last convolutional layer. The resultant representation has a fixed-length of $21\times256=5376$, which is also the fixed input size of the first fully-connected layer (FC$_1$), namely $L_1=pL=5376$.

\begin{table}[htbp]
	\centering\caption{Parameter setting of the convolutional layers in SPP-CNN.}
	\begin{tabular}{|c|c|c|c|c|} \hline
		Group& Layer & {\begin{tabular}[c]{@{}c@{}}Kernel\\size\end{tabular}}& Stride
             & {\begin{tabular}[c]{@{}c@{}}Output\\channel\end{tabular}}\\ \hline
		\multirow{2}{*}{Group 1} & Conv7-64 & $7\times7$ & 1 & 64 \\ \cline{2-5}
		& Max2 & $2\times2$ & 2 & 64 \\ \hline
		\multirow{2}{*}{Group 2} & Conv5-64 & $5\times5$ & 1 & 64 \\ \cline{2-5}
		& Max2 & $2\times2$ & 2 & 64 \\ \hline
		\multirow{2}{*}{Group 3} & Conv3-128 & $3\times3$ & 1 & 128 \\ \cline{2-5}
		& Max2 & $2\times2$ & 2 & 128 \\ \hline
		\multirow{2}{*}{Group 4} & Conv3-128 & $3\times3$ & 1 & 128 \\ \cline{2-5}
		& Max2 & $2\times2$ & 2 & 128 \\ \hline
		\multirow{2}{*}{Group 5} & Conv3-256 & $3\times3$ & 1 & 256 \\ \cline{2-5}
		& Max2 & $2\times2$ & 2 & 256 \\ \hline
		\multirow{2}{*}{Group 6} & Conv3-256 & $3\times3$ & 1 & 256 \\ \cline{2-5}
		& Max2 & $2\times2$ & 2 & 256 \\ \hline
	\end{tabular}
\label{tab:spp_cnn_para}
\end{table}

The parameter setting of the convolutional layers is shown in Table \ref{tab:spp_cnn_para}.
Source codes of this work are available for the public\footnote{\url{https://fylou.github.io/sourcecode.html}}.

\section{Experimental Studies}\label{sec:exp}

In this section, the proposed SPP-CNN is experimentally tested and compared with CNN-RP\cite{Lou2022TCYB}, PATCHY-SAN\cite{Niepert2016ICML}, and LFR-CNN\cite{Lou2022arXiv}. Network connectivity robustness under maximum-degree node attacks and controllability robustness under random node attacks are predicted. Other network robustness under different attack strategies can be studied in the same manner. Both directed and undirected, and synthetic and real-world networks are simulated. All experiments are performed on a PC with GeForce RTX 3080 GPU, which has memory (RAM) 10 GB with running the Ubuntu 20.04.3 LTS Operating System.

This section is organized as follows.
Subsection \ref{exp:set} introduces the experimental settings. General comparison of robustness prediction performances are presented in Subsection \ref{exp:main}, which demonstrates that SPP-CNN performs as well as other state-of-the-art GNN-based robustness predictors when the test data drawn from the same distribution are used for training. The run-time comparison presented in Subsection \ref{exp:rt} demonstrates that SPP-CNN is significantly faster than the GNN-based predictors. Then, the generlizability of the predictors is verified in Subsections \ref{exp:gen} and \ref{exp:real}, where the test datasets with unseen synthetic and real-world networks are tested. Finally, the pros and cons of different predictors are discussed in Subsection \ref{exp:disc}.

\subsection{Experimental Settings}\label{exp:set}

Nine representative synthetic network models are used for simulation, including Barab{\'a}si--Albert (BA) scale-free\cite{Barabasi1999SCI,Barabasi2009SCI}, extreme homogeneous (EH)\cite{Lou2020TCASI}, Erd{\"{o}}s-R{\'e}nyi (ER) random-graph\cite{Erdos1964RG}, \textit{q}-snapback (QS)\cite{Lou2018TCASI}, random hexagon (RH)\cite{Lou2022ACCESS}, random triangle (RT)\cite{Lou2022ACCESS}, generic scale-free (SF)\cite{Goh2001PRL}, Newman--Watts small-world (SW-NW)\cite{Newman1999PLA}, and Watts--Strogatz small-world (SW-WS)\cite{Watts1998N} network models.

Specifically, the generation of BA networks is based on preferential attachment\cite{Barabasi1999SCI}; SF networks are generated using predefined weights, namely the probability of connecting two nodes $i$ and $j$ is proportional to their weights $w_i$ and $w_j$, where $w_{i}=(i+\theta)^{-\sigma}$ for any node $i$, parameters $\sigma\in[0,1)$ and $\theta\ll N$; EH networks are generated by performing random edge rectifications onto ER networks, such that the degree distribution of EH is extremely homogeneous\cite{Lou2020TCASI}; RT and RH consist of random triangles and hexagons, respectively\cite{Lou2022ACCESS}. These models, BA, EH, ER, RH, RT, SF, generate undirected instances by default, while directed instances are generated by assigning random directions onto the edges.

QS consists of a directed backbone chain and multiple snapback edges\cite{Lou2018TCASI}; SW starts from an $N$-node directed loop having $K$ ($K=2$ here) connected nearest-neighbors; shortcuts are randomly added without removing any existing edges in SW-NW\cite{Newman1999PLA}, while rewiring operations are performed in SW-WS\cite{Watts1998N}. Thus, QS, SW-NW, and SW-WS generate directed network instances by default, while undirected instances are generated by removing the edge directions.

The average degree of each network instance is assigned reasonably at random. For directed networks, the average degree range is set as $\langle k\rangle\in[2.5,5]$ for the two SW models, $\langle k\rangle\in[2,4]$ for RH, $\langle k\rangle\in[1.5,3]$ for RT, and for the other models, $\langle k\rangle\in[3,6]$. For undirected networks, the average degree range is set to double, namely, the range is set as $\langle k\rangle\in[5,10]$ for the two SW models, $\langle k\rangle\in[4,8]$ for RH, $\langle k\rangle\in[3,6]$ for RT, while for the other models, $\langle k\rangle\in[6,12]$.

Define three sets of synthetic network models, $S_{1}=$\{BA, EH, ER, QS, RH, RT, SF, SW-NW, SW-WS\}, $S_{2}=$\{ER, QS, SF, SW-NW\}, and $S_{3}=$\{BA, EH, RH, RT, SW-WS\}. Three network size ranges are set as $N_{a}\in[700,1300]$, $N_{b}\in[300,700]$, $N_{c}\in[1300,1700]$. For each synthetic model, 1000 network instances are randomly generated as the training data, and 100 instances as the test data. Since the structural connectivity and controllability are independent of the edge weights, only unweighted networks are simulated here.

For the LFR module in both PATCHY-SAN and LFR-CNN, the number of features is set as $h=2$, where node degree and clustering coefficient are employed; the size of receptive field is set as $g=10$, both are the same as that in\cite{Niepert2016ICML,Lou2022arXiv}.

\subsection{Comparison of Prediction Performances}\label{exp:main}

Both training and test data are drawn from the same sample space, namely each network instance is randomly generated from any of the 9 synthetic models in $S_{1}$, with a network size randomly picked as $N_{a}\in[700,1300]$.

\begin{table*}[htbp]
	\centering\caption{Comparison of average prediction errors among SPP-CNN, CNN-RP, PATCHY-SAN, and LFR-CNN. Signs in parentheses denote the Kruskal-Wallis H-test \cite{Kruskal1952JASA} results. A `$+$' sign denotes that SPP-CNN is superior to the other methods with smaller errors; a `$\approx$' sign denotes no significant difference; while a `$-$' sign denotes that SPP-CNN is inferior to the other methods with greater errors.}
	\begin{tabular}{|ccl|l|l|l|l|l|l|l|l|l|}
		\hline
		\multicolumn{3}{|c|}{} & \multicolumn{1}{c|}{BA} & \multicolumn{1}{c|}{EH} & \multicolumn{1}{c|}{ER} & \multicolumn{1}{c|}{QS} & \multicolumn{1}{c|}{RH} & \multicolumn{1}{c|}{RT} & \multicolumn{1}{c|}{SF} & \multicolumn{1}{c|}{\begin{tabular}[c]{@{}c@{}}SW-\\ NW\end{tabular}} & \multicolumn{1}{c|}{\begin{tabular}[c]{@{}c@{}}SW-\\ WS\end{tabular}} \\ \hline
		\multicolumn{1}{|c|}{\multirow{8}{*}{\rotatebox[origin=c]{90}{Directed}}} & \multicolumn{1}{c|}{\multirow{4}{*}{{\begin{tabular}[c]{@{}c@{}}Connectivity\\Robustness\\Eq. (\ref{eq:nlc})\end{tabular}}}} & SPP-CNN & 0.038 & 0.057 & 0.032 & 0.067 & 0.056 & 0.105 & 0.017 & 0.030 & 0.031 \\ \cline{3-12}
		\multicolumn{1}{|c|}{} & \multicolumn{1}{c|}{} & CNN-RP & 0.146($+$) & 0.138($+$) & 0.129($+$) & 0.162($+$) & 0.121($+$) & 0.118($\approx$) & 0.068($+$) & 0.094($+$) & 0.099($+$) \\ \cline{3-12}
		\multicolumn{1}{|c|}{} & \multicolumn{1}{c|}{} & PATCHY-SAN & 0.040($+$) & 0.032($-$) & 0.024($-$) & 0.027($-$) & 0.028($-$) & 0.031($-$) & 0.029($+$) & 0.027($\approx$) & 0.024($-$) \\ \cline{3-12}
		\multicolumn{1}{|c|}{} & \multicolumn{1}{c|}{} & LFR-CNN & 0.041($+$) & 0.078($+$) & 0.039($\approx$) & 0.076($\approx$) & 0.023($-$) & 0.031($-$) & 0.018($\approx$) & 0.029($\approx$) & 0.027($\approx$) \\ \cline{2-12}
		\multicolumn{1}{|c|}{} & \multicolumn{1}{c|}{\multirow{4}{*}{{\begin{tabular}[c]{@{}c@{}}Controllability\\Robustness\\Eq. (\ref{eq:nd})\end{tabular}}}} & SPP-CNN & 0.031 & 0.025 & 0.025 & 0.042 & 0.021 & 0.030 & 0.025 & 0.028 & 0.025 \\ \cline{3-12}
		\multicolumn{1}{|c|}{} & \multicolumn{1}{c|}{} & CNN-RP & 0.092($+$) & 0.040($+$) & 0.062($+$) & 0.104($+$) & 0.051($+$) & 0.060($+$) & 0.134($+$) & 0.054($+$) & 0.059($+$) \\ \cline{3-12}
		\multicolumn{1}{|c|}{} & \multicolumn{1}{c|}{} & PATCHY-SAN & 0.032($\approx$) & 0.028($\approx$) & 0.026($\approx$) & 0.030($-$) & 0.019($\approx$) & 0.024($\approx$) & 0.049($+$) & 0.024($\approx$) & 0.022($\approx$) \\ \cline{3-12}
		\multicolumn{1}{|c|}{} & \multicolumn{1}{c|}{} & LFR-CNN & 0.024($\approx$) & 0.013($-$) & 0.018($-$) & 0.015($-$) & 0.015($-$) & 0.019($-$) & 0.039($+$) & 0.015($-$) & 0.015($-$) \\ \hline
		\multicolumn{1}{|c|}{\multirow{8}{*}{\rotatebox[origin=c]{90}{Undirected}}} & \multicolumn{1}{c|}{\multirow{4}{*}{{\begin{tabular}[c]{@{}c@{}}Connectivity\\Robustness\\Eq .(\ref{eq:nlc})\end{tabular}}}} & SPP-CNN & 0.023 & 0.070 & 0.025 & 0.026 & 0.028 & 0.047 & 0.016 & 0.029 & 0.028 \\ \cline{3-12}
		\multicolumn{1}{|c|}{} & \multicolumn{1}{c|}{} & CNN-RP & 0.055($+$) & 0.188($+$) & 0.121($+$) & 0.135($+$) & 0.121($+$) & 0.100($+$) & 0.020($\approx$) & 0.149($+$) & 0.140($+$) \\ \cline{3-12}
		\multicolumn{1}{|c|}{} & \multicolumn{1}{c|}{} & PATCHY-SAN & 0.030($+$) & 0.032($-$) & 0.022($-$) & 0.024($\approx$) & 0.022($-$) & 0.026($-$) & 0.025($+$) & 0.025($\approx$) & 0.020($-$) \\ \cline{3-12}
		\multicolumn{1}{|c|}{} & \multicolumn{1}{c|}{} & LFR-CNN & 0.026($\approx$) & 0.066($\approx$) & 0.025($\approx$) & 0.027($\approx$) & 0.019($-$) & 0.022($-$) & 0.016($\approx$) & 0.023($\approx$) & 0.020($-$) \\ \cline{2-12}
		\multicolumn{1}{|c|}{} & \multicolumn{1}{c|}{\multirow{4}{*}{{\begin{tabular}[c]{@{}c@{}}Controllability\\Robustness\\Eq. (\ref{eq:nd})\end{tabular}}}} & SPP-CNN & 0.027 & 0.013 & 0.016 & 0.014 & 0.018 & 0.018 & 0.026 & 0.016 & 0.017 \\ \cline{3-12}
		\multicolumn{1}{|c|}{} & \multicolumn{1}{c|}{} & CNN-RP & 0.060($+$) & 0.037($+$) & 0.031($+$) & 0.035($+$) & 0.039($+$) & 0.039($+$) & 0.085($+$) & 0.045($+$) & 0.046($+$) \\ \cline{3-12}
		\multicolumn{1}{|c|}{} & \multicolumn{1}{c|}{} & PATCHY-SAN & 0.028($\approx$) & 0.015($\approx$) & 0.020($\approx$) & 0.017($+$) & 0.021($+$) & 0.020($\approx$) & 0.040($+$) & 0.016($\approx$) & 0.016($\approx$) \\ \cline{3-12}
		\multicolumn{1}{|c|}{} & \multicolumn{1}{c|}{} & LFR-CNN & 0.037($+$) & 0.016($+$) & 0.017($\approx$) & 0.017($+$) & 0.018($\approx$) & 0.020($\approx$) & 0.060($+$) & 0.018($+$) & 0.019($\approx$) \\ \hline
	\end{tabular}\label{tab:gen}
\end{table*}

Table \ref{tab:gen} shows the average prediction errors obtained by SPP-CNN, CNN-RP, PATCHY-SAN, and LFR-CNN. Boxplots of the prediction errors are presented in Figs. S1--S4 of the Supplementary Information (SI)\footnote{\url{https://fylou.github.io/pdf/sppsi.pdf}} due to space limitation here. Network robustness in terms of connectivity and controllability is predicted, and the prediction errors are calculated using Eq. (\ref{eq:avg_xi}). The errors are averaged from 100 independent runs for each synthetic model. A total of $2\times2\times9=36$ comparisons are performed between SPP-CNN and each of CNN-RP, PATCHY-SAN, and LFR-CNN, namely, directed and undirected networks, two robustness measures, and 9 synthetic network models. The Kruskal-Wallis H-test\cite{Kruskal1952JASA} results are shown with the corresponding prediction error values, where a `$+$' sign denotes that SPP-CNN performs significantly better than the other methods with smaller errors; a `$\approx$' sign denotes no significant difference between SPP-CNN and the other methods; while a `$-$' sign denotes that SPP-CNN performs significantly worse than the other methods with greater errors.

SPP-CNN performs significantly better than CNN-RP in 34 cases, and for the rest 2 cases, there is no significant difference between them. As for PATCHY-SAN and LFR-CNN, the same numbers of superiors and inferiors are obtained, showing that SPP-CNN significantly outperforms PATCHY-SAN (or LFR-CNN) in 8 cases; SPP-CNN performs significantly worse than PATCHY-SAN (or LFR-CNN) in 12 cases; and they perform statistically equally in the rest 16 cases.

In a nutshell, it is clear that SPP-CNN performs equivalently to or marginally worse than PATCHY-SAN and LFR-CNN, but significantly better than CNN-RP.

\subsection{Run Time Comparison} \label{exp:rt}

The powerful GNN-based representation learning part of PATCHY-SAN and LFR-CNN significantly enhances the precision of robustness prediction, which however is the most time-consuming part. In contrast, SPP-CNN does not involve any GNN-based operation for feature learning.

\begin{figure}[htbp]
	\begin{subfigure}{.25\textwidth}
		\centering\includegraphics[width=\linewidth]{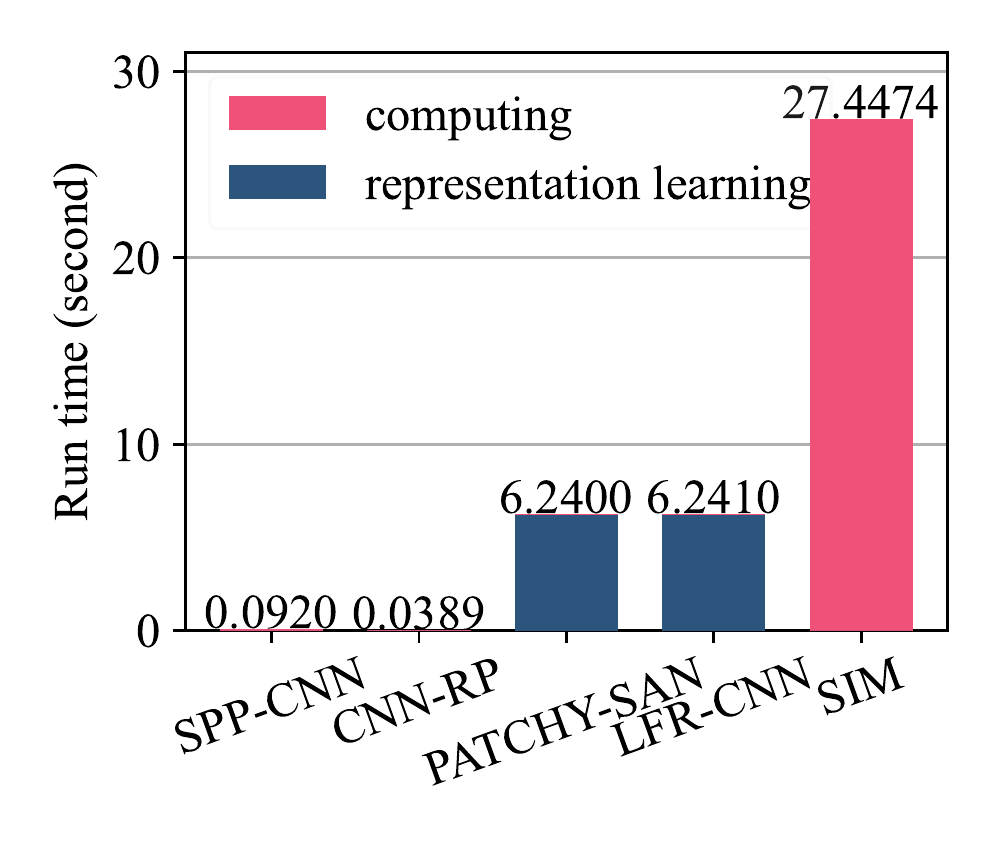}
		\caption{$N_{a}\in[700,1300]$}
		\label{fig:t713}
	\end{subfigure}%
	\begin{subfigure}{.25\textwidth}
		\centering\includegraphics[width=\linewidth]{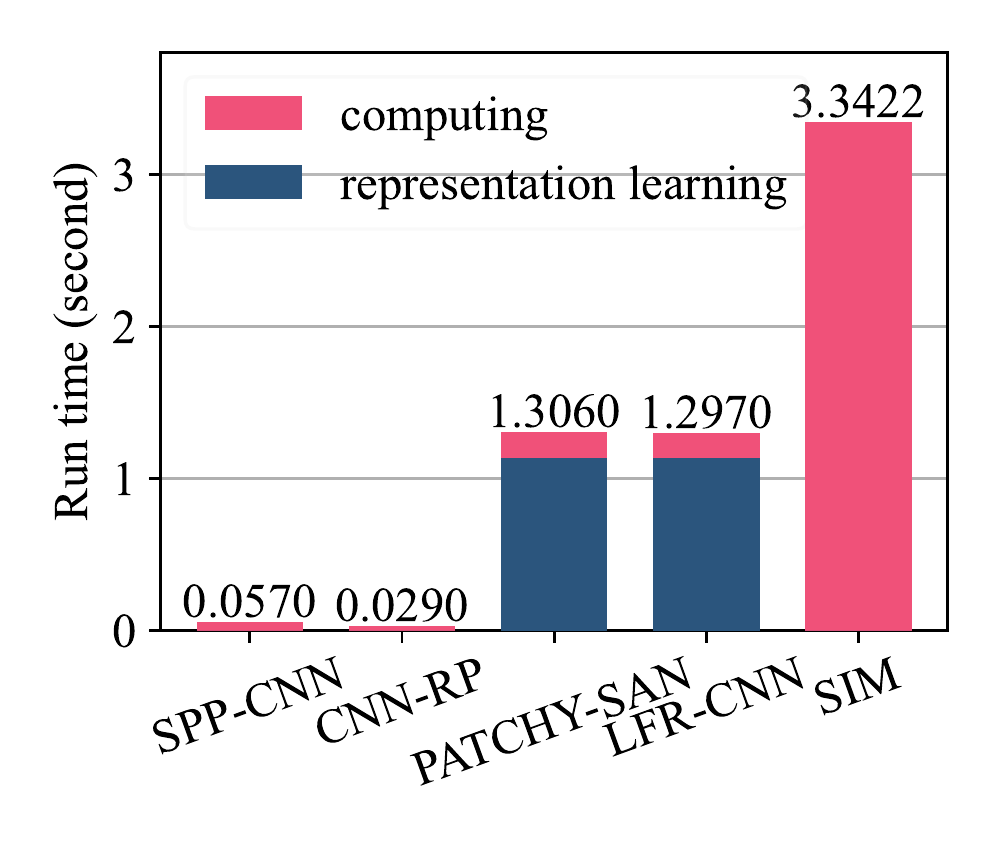}
		\caption{$N_{b}\in[300,700]$}
		\label{fig:t37}
	\end{subfigure}%
	\caption{Run time comparison of SPP-CNN, CNN-RP, PATCHY-SAN, LFR-CNN, and attack simulation (SIM): (a) for network size $N_{a}\in[700,1300]$; and (b) for network size $N_{b}\in[300,700]$ }
	\label{fig:fig_im}
\end{figure}

Figure \ref{fig:fig_im} shows the run time comparison of SPP-CNN, CNN-RP, PATCHY-SAN, LFR-CNN, and attack simulation (SIM). Fig. \ref{fig:fig_im} (a) presents the run time when the network size is $N_{a}\in[700,1300]$, of which the average network size is 1000;
for Fig. \ref{fig:fig_im} (b), the network size is set as $N_{b}\in[700,1300]$, of which the average network size is 500.

For each subplot in Fig. \ref{fig:fig_im}, it is clear that CNN-RP is the fastest, followed by SPP-CNN. Clearly, PATCHY-SAN and LFR-CNN are significantly slower than CNN-RP and SPP-CNN, but much faster than attack simulations. For PATCHY-SAN and LFR-CNN, the representation learning takes most of the run time.

Comparing Figs. \ref{fig:fig_im} (a) and (b), when the average network size is doubled from subplot (b) to subplot (a), the time consumption of SPP-CNN, CNN-RP, PATCHY-SAN, LFR-CNN, and SIM is increased by 1.61, 1,34, 4.78, 4.81, and 8.21 times, respectively. It is clear that the time consumption of CNN-RP and SPP-CNN increase significantly slower than the GNN-based predictors do.

The run time difference between CNN-RP and SPP-CNN is negligible, while SPP-CNN significantly outperforms CNN-RP, as can be seen from Table \ref{tab:gen}, which also shows that the performance different between SPP-CNN and the GNN-based predictors is marginal, while the time complexity of SPP-CNN is significantly lower than PATCHY-SAN and LFR-CNN, as shown in Fig. \ref{fig:fig_im}.

\subsection{Comparison of Generalizability} \label{exp:gen}

The generalizability of robustness predictors is tested from two aspects: 1) using the network models in $S_{1}$, two sets of test instances with unseen network sizes (UNS) are tested, namely, the network size of training data is drawn from $N_{a}\in[700,1300]$, while the test data are from $N_{b}\in[300,700]$ and $N_{c}\in[1300,1700]$.
2) Given network size drawn from $N_{a}\in[700,1300]$, training instances are generated from the network models in $S_{2}$, while the test instances are from $S_{3}$. Since $S_{2}$ and $S_{3}$ are two mutually exclusive subsets of $S_{1}$, $S_{3}$ is called the test data of unseen network topology (UNT).  Note that although $S_{2}$ and $S_{3}$ are mutually exclusive, there are some similarities between the models in the two sets, e.g., SW-NW $\in S_{2}$ and SW-WS $\in S_{3}$.

\begin{table}[htbp]
	\centering\caption{Comparison of significant difference between SPP-CNN and each of CNN-RP, PATCHY-SAN, and LFR-CNN. }
	\begin{tabular}{|cl|c|c|c|}
		\hline
		\multicolumn{2}{|c|}{Significant Difference} & ($+$) & ($-$) & ($\approx$) \\ \hline
		\multicolumn{1}{|c|}{\multirow{3}{*}{UNS}} & SPP-CNN vs CNN-RP & 266 & 12 & 10 \\ \cline{2-5}
		\multicolumn{1}{|c|}{} & SPP-CNN vs PATCHY-SAN & 169 & 78 & 41 \\ \cline{2-5}
		\multicolumn{1}{|c|}{} & SPP-CNN vs LFR-CNN & 124 & 108 & 56 \\ \hline
		\multicolumn{1}{|c|}{\multirow{3}{*}{UNT}} & SPP-CNN vs CNN-RP & 11 & 2 & 7 \\ \cline{2-5}
		\multicolumn{1}{|c|}{} & SPP-CNN vs PATCHY-SAN & 15 & 3 & 2 \\ \cline{2-5}
		\multicolumn{1}{|c|}{} & SPP-CNN vs LFR-CNN & 14 & 3 & 3 \\ \hline
	\end{tabular}\label{tab:un}
\end{table}

Table \ref{tab:un} summarizes the comparison of the significant differences between SPP-CNN and each of CNN-RP, PATCHY-SAN, and LFR-CNN.  The prediction errors for UNS and UNT are presented in Figs. S5--S8 and S13 of SI, respectively. The barcharts of the numbers of significant performance differences for UNS and UNT are shown in Figs. S9--S12 and S14 of SI, respectively.
For UNS, there are $2\times2\times9\times8=288$ neck-to-neck comparisons between SPP-CNN and each of CNN-RP, PATCHY-SAN, and LFR-CNN, namely, directed and undirected, connectivity and controllability robustness, 9 synthetic models in $S_1$, and 8 UNS sections in $N_{b}$ and $N_{c}$.
For UNT, there are $2\times2\times5=20$ comparisons, namely, directed and undirected, connectivity and controllability robustness, and 5 synthetic models in $S_3$.

Only when testing on UNS, SPP-CNN outperforms LFR-CNN marginally, but for all the rest comparisons, SPP-CNN gains more superiors than inferiors. This indicates the excellent generalizability of SPP-CNN.

\subsection{Predicting Robustness for Real-world Networks} \label{exp:real}

The prediction performance is studied by two experiments. The first experiment investigates the performances of SPP-CNN, PATCHY-SAN, and LFR-CNN on predicting the robustness of the tested real-world networks. The second experiment uses mixed sets of both synthetic and real-world networks as both training and test data. Define $S_{r}=$\{Reddit-Multi-12K\footnote{\url{https://networkrepository.com/REDDIT-MULTI-12K.php}} Networks\}, where there are 1000 real-world networks selected randomly, of which the network size ranges as $N_{r}\in[300,700]$.

\begin{figure*}[htbp]
 	\centering
 	\begin{subfigure}{.30\textwidth}
 		\centering\includegraphics[width=\linewidth]{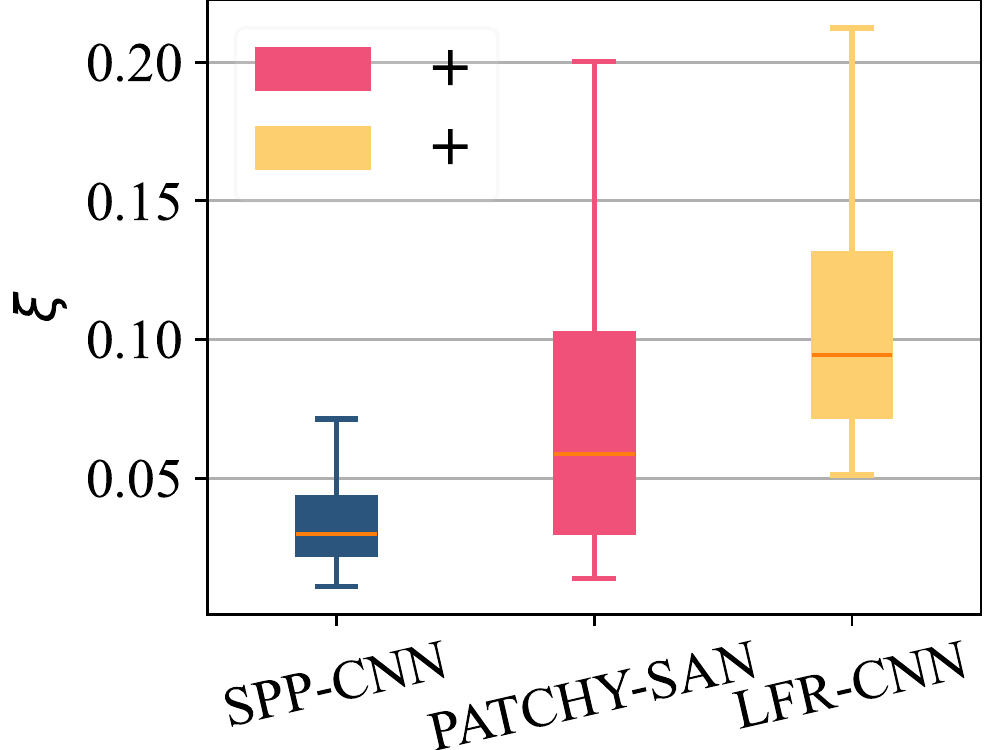}
 		\caption{Real-world networks} \label{fig:e_real}
 	\end{subfigure}
 	\begin{subfigure}{.30\textwidth}
	 	\centering\includegraphics[width=\linewidth]{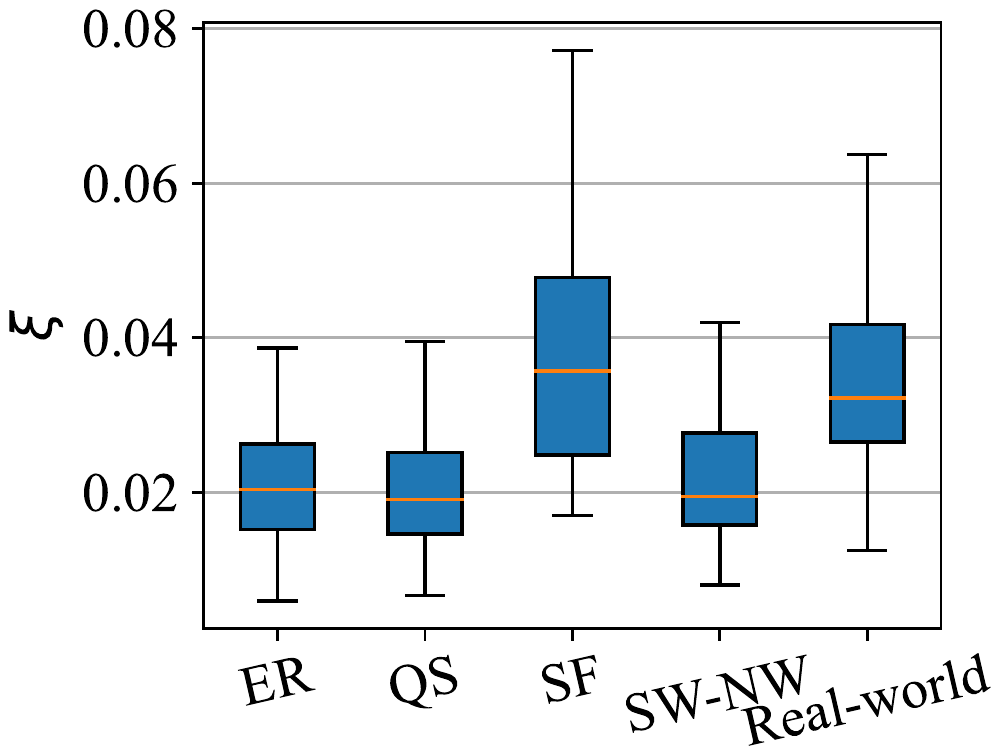}
	 	\caption{Real-world networks and $S_{2}$} \label{fig:e_mix}
	\end{subfigure}
	\begin{subfigure}{.30\textwidth}
		\centering\includegraphics[width=\linewidth]{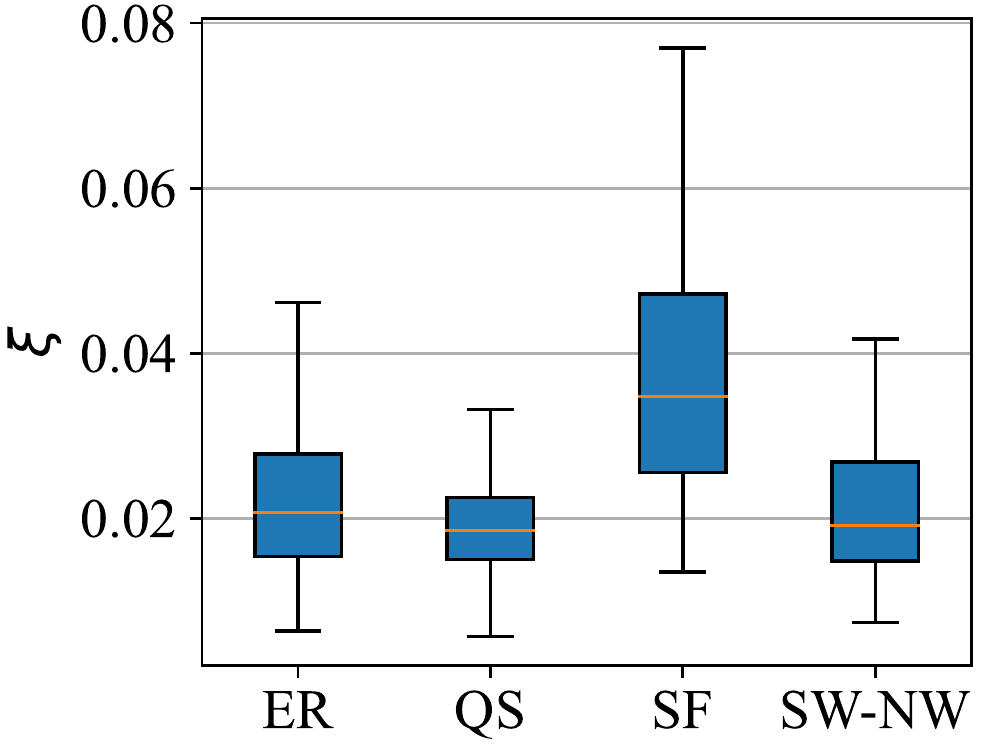}
		\caption{$S_{2}$ only} \label{fig:e_syn}
	\end{subfigure}%
	\caption{Prediction error comparison in the form of boxplot: (a) both training and test data are real-world networks; (b) for SPP-CNN, the training and test data include both real-world networks $S_{r}$ and synthetic models $S_2$; (c) the benchmark performance of SPP-CNN when only $S_2$ is used as both training and test data. }\label{fig:real}
\end{figure*}

\subsubsection{Real-world Networks Only}

Figure \ref{fig:real} (a) shows the prediction errors obtained by SPP-CNN, PATCHY-SAN, and LFR-CNN, where the training data set is $S_{r}$ while the test set constitutes of other randomly picked 100 networks from Reddit-Multi-12K. The boxplot in Fig. \ref{fig:real} (a) shows that SPP-CNN outperforms both PATCHY-SAN and LFR-CNN, with statistic significance, using the Kruskal-Wallis test\cite{Kruskal1952JASA}.

\subsubsection{Real-world and Synthetic Networks}

To clear up the doubt that the excellent performance of SPP-CNN on predicting real-world networks is random or due to overfitting, $S_{r}$ and $S_2$ are mixed together in experiments. Thus, the trained SPP-CNN is neither specialized for synthetic models nor for real-world networks.

As shown in Fig. \ref{fig:real} (b), the prediction error boxplot obtained by SPP-CNN on real-world networks is similar to that in Fig. \ref{fig:real} (a). This clears up the doubt of overfitting. Also, the prediction errors obtained by SPP-CNN for synthetic networks are similar to that in Fig. \ref{fig:real} (c), where a SPP-CNN is specifically trained for the 4 synthetic models.

In a nutshell, when real-world and synthetic networks are mixed and used as both training and test data, SPP-CNN can perform robustness prediction as excellent as the cases where only real-world or only synthetic networks are used. This also implies the excellent generalizability of SPP-CNN.

\subsection{Discussions} \label{exp:disc}

\subsubsection{Overall Performance}

\begin{table*}[htbp]
	\centering\caption{A summary of the performances of SPP-CNN when different datasets (including $S_1$, $S_2$, $S_3$, and $S_r$) and network sizes (including $N_a$, $N_b$, and $N_c$) are used. A `$\approx$' sign represents that the numbers of superiors and inferiors obtained by SPP-CNN are similar to that obtained by the compared method; while a `$\succ$' sign means that SPP-CNN obtains clearly more superiors than inferiors in the corresponding comparison. }
	\begin{tabular}{|c|c|c|c|c|c|}
		\hline
		\begin{tabular}[c]{@{}c@{}}Training data\end{tabular} & $S_1$ ($N_a$) & $S_1$ ($N_a$) & $S_2$ ($N_a$) & $S_r$ & $S_r$+$S_2$ \\ \hline
		\begin{tabular}[c]{@{}c@{}}Test data\end{tabular} & $S_1$ ($N_a$) & $S_1$ ($N_b$ and $N_c$) & $S_3$ ($N_a$) & $S_r$ & $S_r$+$S_2$ \\ \hline
		\begin{tabular}[c]{@{}c@{}}Performance\\ of SPP-CNN\end{tabular} & \multicolumn{1}{l|}{\begin{tabular}[c]{@{}l@{}}$\approx$ PATCHY-SAN\\$\approx$ LFR-CNN\\$\succ$CNN-RP\end{tabular}} & \multicolumn{1}{l|}{\begin{tabular}[c]{@{}l@{}}$\succ$ PATCHY-SAN\\$\approx$ LFR-CNN\\$\succ$ CNN-RP\end{tabular}} & \multicolumn{1}{l|}{\begin{tabular}[c]{@{}l@{}}$\succ$ PATCHY-SAN\\$\succ$ LFR-CNN\\$\succ$ CNN-RP\end{tabular}} & \multicolumn{1}{l|}{\begin{tabular}[c]{@{}l@{}}$\succ$ PATCHY-SAN\\$\succ$ LFR-CNN\\$\succ$ CNN-RP\end{tabular}} &
		\multicolumn{1}{l|}{\begin{tabular}[c]{@{}l@{}}as good as when $S_r$\\and $S_2$ are trained\\and tested separately\end{tabular}}\\ \hline
	\end{tabular}\label{tab:overview}
\end{table*}
\begin{figure}[htbp]
	\centering \includegraphics[width=\linewidth]{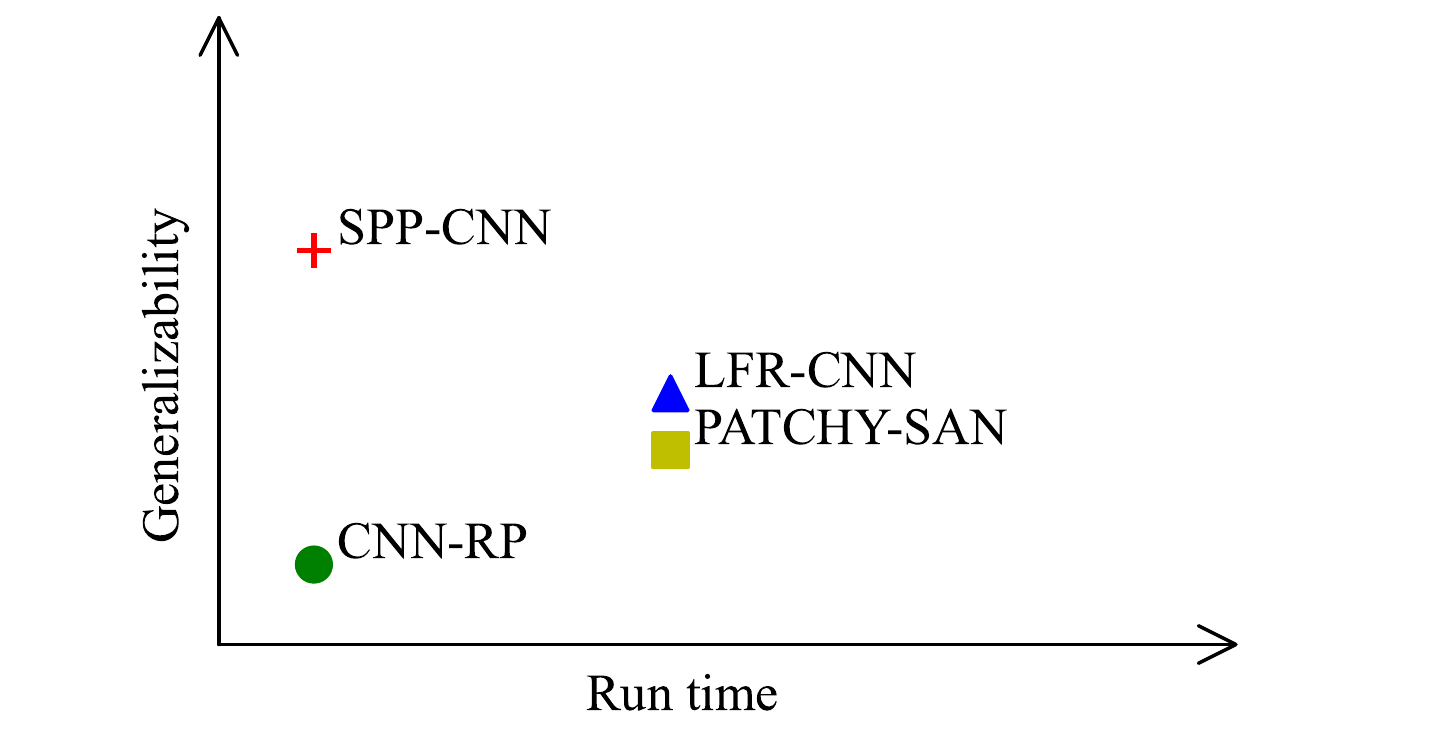}
	\caption{Overall performances of SPP-CNN, CNN-RP, PATCHY-SAN, and LFR-CNN. }\label{fig:overview}
\end{figure}

Table \ref{tab:overview} shows the overall performances of SPP-CNN compared to CNN-RP, PATCHY-SAN, and LFR-CNN. The results are summarized from the results presented in Subsections \ref{exp:main}, \ref{exp:gen}, and \ref{exp:real}. When the test data and training data are drawn from the same sample space, SPP-CNN performs as good as the GNN-based predictors. However, if the test data are drawn from different sample spaces, SPP-CNN consistently outperforms the other predictors. Here, different sample spaces mean that either UNS or UNT is set. Also, SPP-CNN outperforms the GNN-based predictors on predicting the robustness of real-world networks.

Considering both generalizability and run time, Fig. \ref{fig:overview} shows the performance comparison in two-dimensional coordinate plots. Clearly, SPP-CNN and CNN-RP are faster, while SPP-CNN meanwhile possesses the best generalizability.

\subsubsection{Information Loss}

Given an $N\times N$ adjacency matrix as input, CNN-RP requires a fixed input size $W\times W$, and thus downsampling or upsampling is necessary. If $N>W$, then $N-W$ columns and rows are needed to be randomly deleted from the adjacency matrix in order to fit the input size. If $N<W$, then $W-N$ empty columns and rows are needed to be randomly added for padding. These deletion or addition operations may significantly distort the original network topology, and thus degenerate the subsequent robustness prediction performance.

For PATCHY-SAN and LFR-CNN, as discussed in Subsection \ref{subsub:lfr}, if $N>W$, then only the neighboring fields of the $W$ most important nodes are selected to construct the receptive fields. While if $N<W$ then $W-N$ dummy nodes are generated. For CNN-RP, there is always a proportion of $\delta=\frac{|N-W|}{N}$ information distortion, while for PATCHY-SAN and LFR-CNN, this information distortion is significantly lower than $\sigma$, since the neighboring fields of the $W$ most important nodes are always included, but other unimportant nodes may also be compressed into the neighboring fields.

Finally, for SPP-CNN, there is always no information loss or distortion from the input, since networks of any sizes are able to be input without the need of resizing. Moreover, local features are better captured by the SPP layer.

\section{Conclusions}\label{sec:end}

Measuring network robustness by attack simulations is time-consuming, while deep neural networks provide a more cost-effective technique for robustness prediction, which can replace the iterative attack simulations, at least partially. The CNN-based framework CNN-RP can predict network robustness fast, but is inefficient when the size of the concerned network is different from the fixed input size of the CNN. On the other hand, PATCHY-SAN and LFR-CNN, which incorporate both GNN and CNN, are able to predict network robustness of different sizes and various topologies with low prediction errors. However, these GNN-based robustness predictors perform significantly slower than CNN-RP, due to the powerful but time-consuming feature learning module installed.

In this paper, to overcome the mismatch issue between the various network sizes and the somewhat fixed input size of the CNN, a spatial pyramid pooling layer is installed between the convolutional and fully-connected layers of the CNN, yielding the new SPP-CNN framework.

Extensive experiments are carried out by comparing SPP-CNN with CNN-RP, PATCHY-SAN, and LFR-CNN. Three sets of synthetic (directed and undirected) networks with three different network size ranges, together with one set of real-world networks, are simulated. Detailed comparisons are performed, where both training and test data are drawn from the same sample space. Generalization abilities of the predictors are also examined, where the test data are drawn from different sample spaces, other than the training data space. The prediction performances on real-world networks are tested from two aspects: 1) real-world networks are trained and tested separately, and 2) real-world networks are mixed with synthetic networks. All the experimental results demonstrate the excellent performances of SPP-CNN: 1) SPP-CNN achieves significantly better prediction performances than CNN-RP with similar performances as PATCHY-SAN and LFR-CNN, when tboth training and test data are drawn from the same sample space. 2) When the sample spaces of training and test data are different, SPP-CNN shows stronger generalizability than the other three predictors. 3) SPP-CNN performs prediction significantly faster than PATCHY-SAN and LFR-CNN.

Overall, the proposed SPP-CNN framework lifts the network robustness prediction to a higher level, so that the prediction tasks can be accomplished faster and more precise. This investigation reveals that the great potential of deep neural networks can be further explored for broader applications in the future.

\bibliographystyle{IEEEtran}
\bibliography{ref}



\section*{Supplementary material}

# Supplementary Information for "*SPP-CNN: An Efficient Framework for Network Robustness Prediction*"

Chengpei Wu, Yang Lou, Lin Wang, Junli Li, Xiang Li and Guanrong Chen

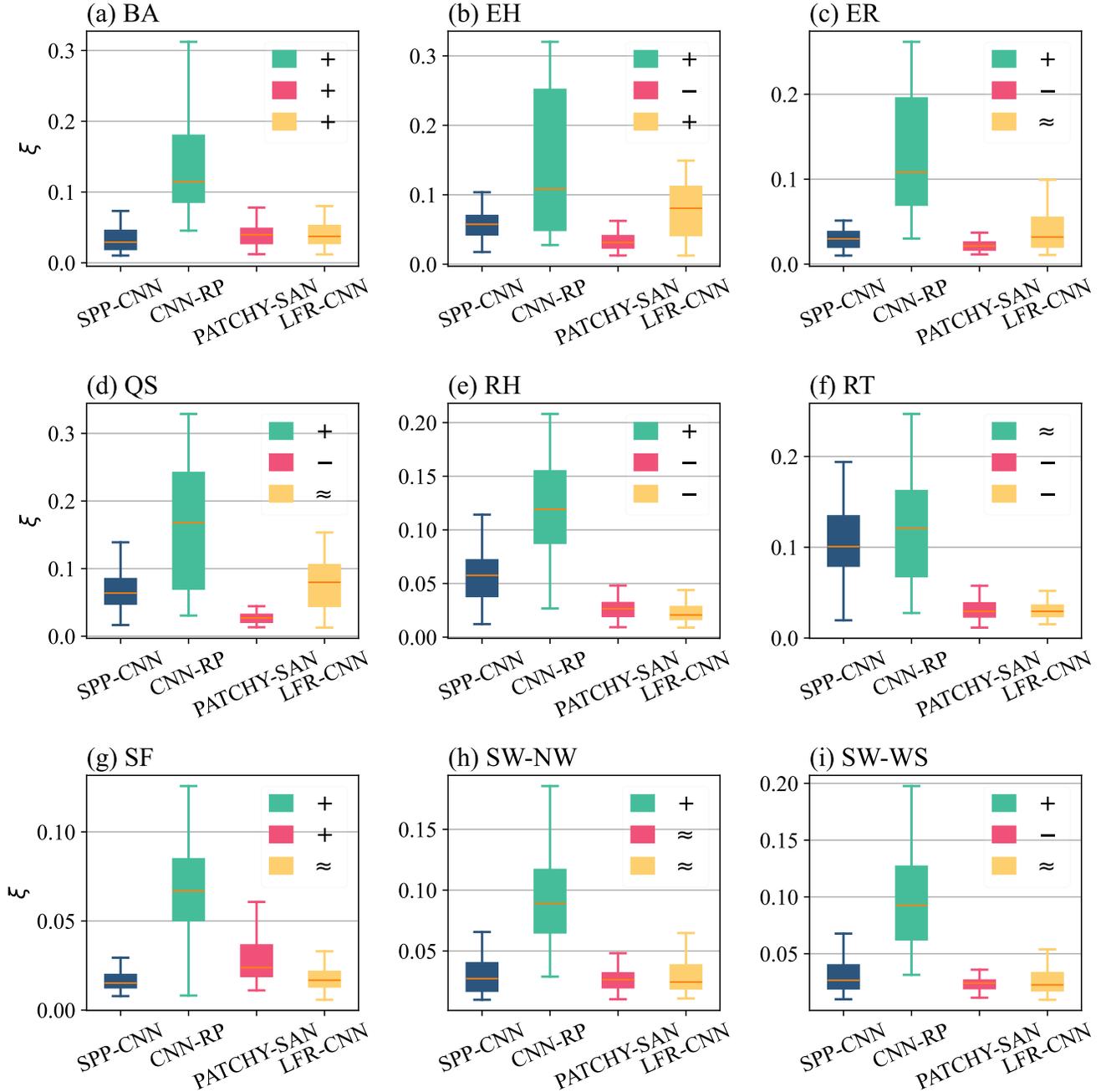

Fig. S1: Boxplots of prediction errors obtained by SPP-CNN, CNN-RP, PATCHY-SAN, and LFR-CNN. Networks of $S_1$ and $N_\alpha \in [700, 1300]$ are used for both training and test datasets. Connectivity robustness of directed networks under maximum-degree node attacks is predicted.

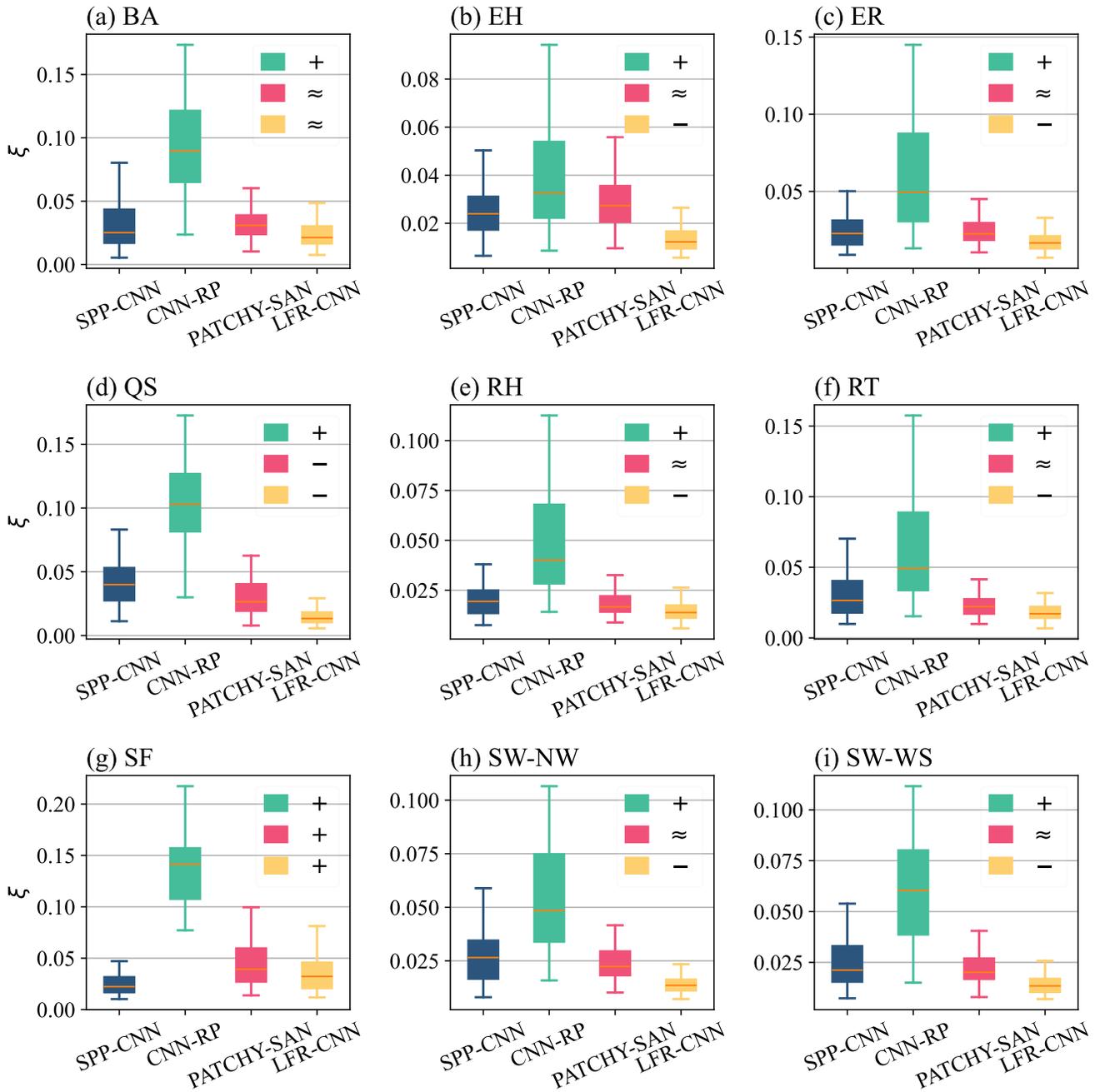

Fig. S2: Boxplots of prediction errors obtained by SPP-CNN, CNN-RP, PATCHY-SAN, and LFR-CNN. Networks of $S_1$ and $N_a \in [700, 1300]$ are used for both training and test datasets. Controllability robustness of directed networks under random node attacks is predicted.

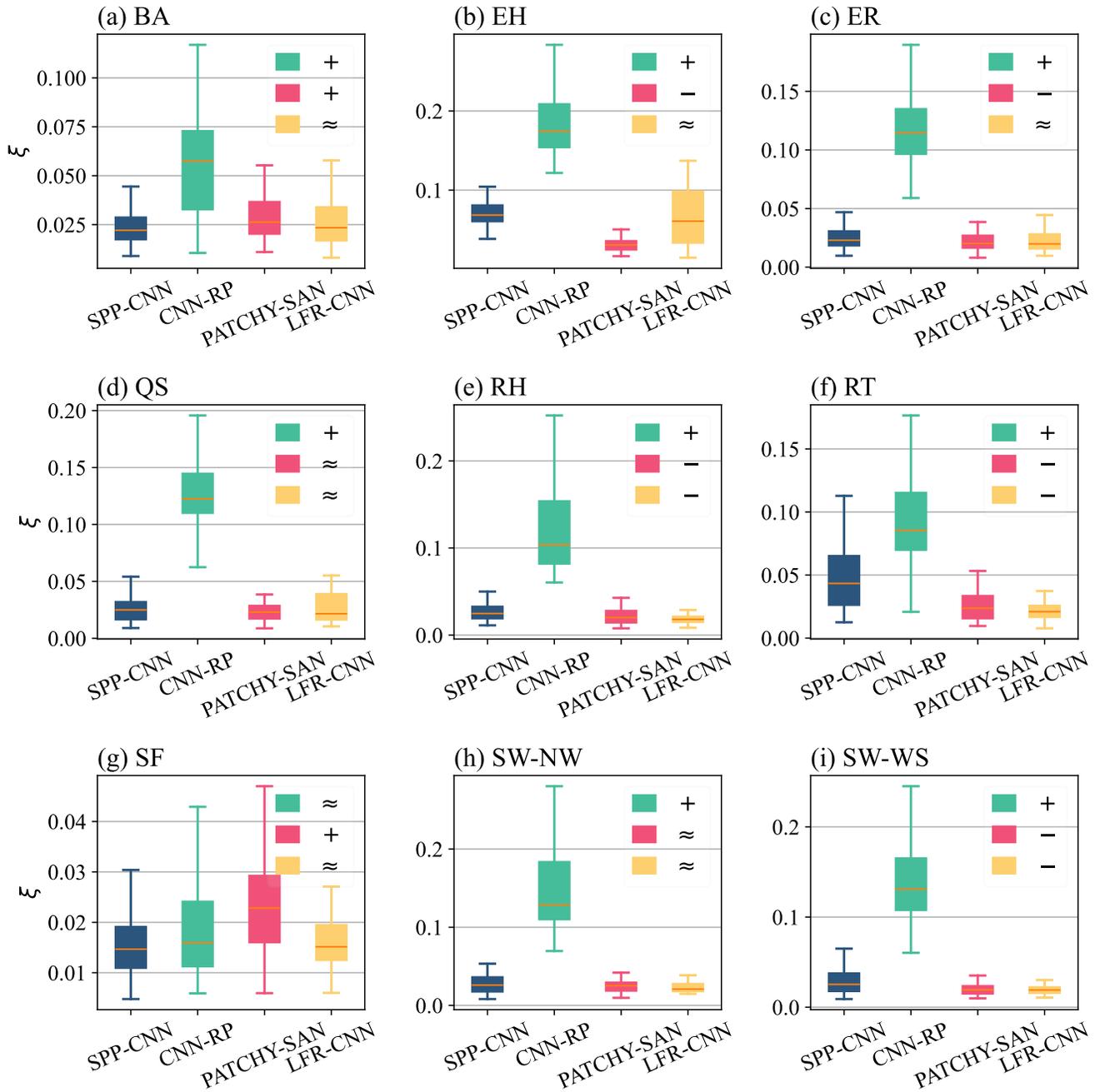

Fig. S3: Boxplots of prediction errors obtained by SPP-CNN, CNN-RP, PATCHY-SAN, and LFR-CNN. Networks of $S_1$ and $N_a \in [700, 1300]$ are used for both training and test datasets. Connectivity robustness of undirected networks under maximum-degree node attacks is predicted.

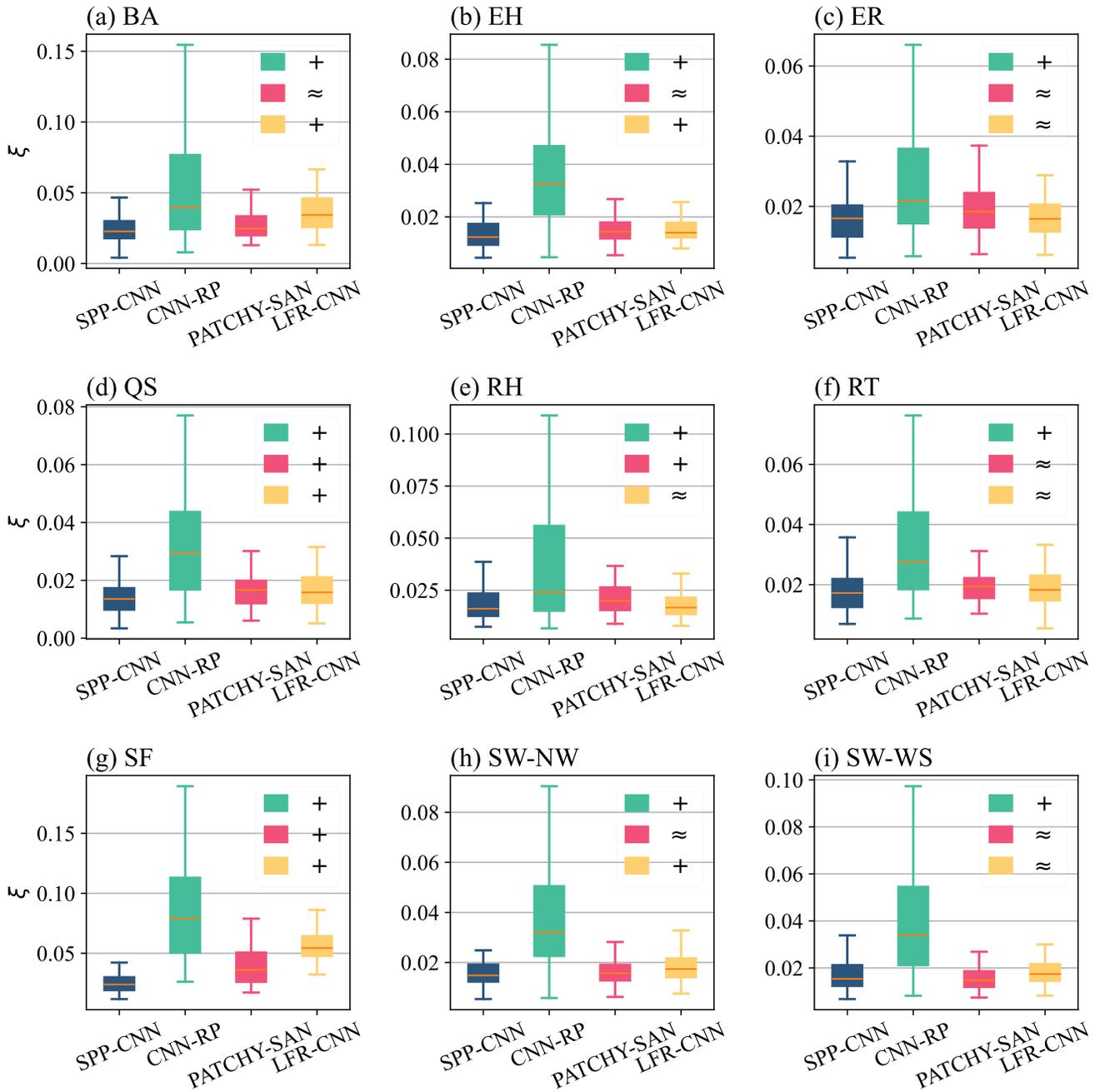

Fig. S4: Boxplots of prediction errors obtained by SPP-CNN, CNN-RP, PATCHY-SAN, and LFR-CNN. Networks of $S_1$ and $N_a \in [700, 1300]$ are used for both training and test datasets. Controllability robustness of undirected networks under random node attacks is predicted.

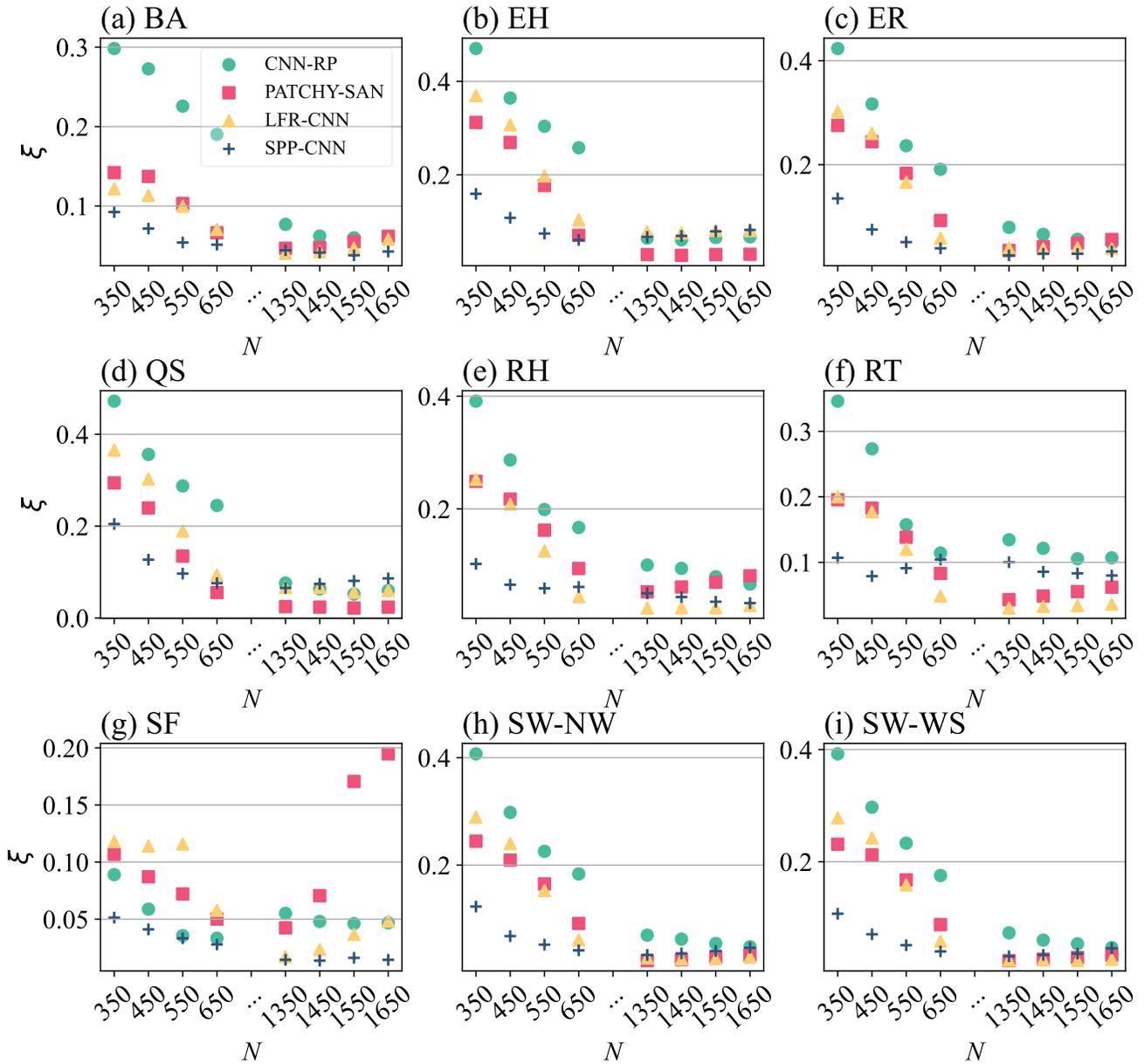

Fig. S5: Prediction errors obtained by SPP-CNN, CNN-RP, PATCHY-SAN, and LFR-CNN for unseen network sizes (UNS). Connectivity robustness of directed networks under maximum-degree node attacks is predicted.

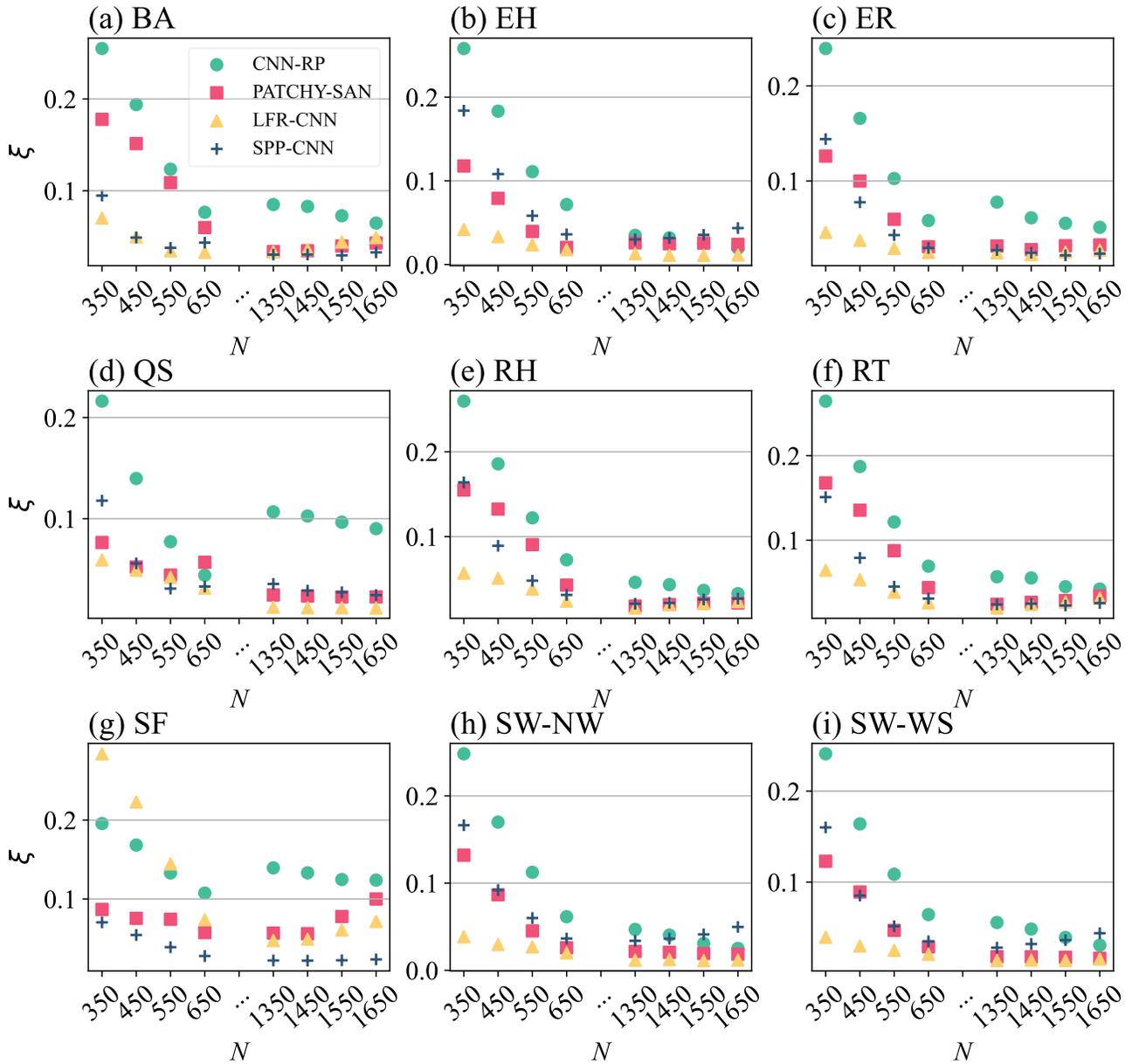

Fig. S6: Prediction errors obtained by SPP-CNN, CNN-RP, PATCHY-SAN, and LFR-CNN for unseen network sizes (UNS). Controllability robustness of directed networks under random node attacks is predicted.

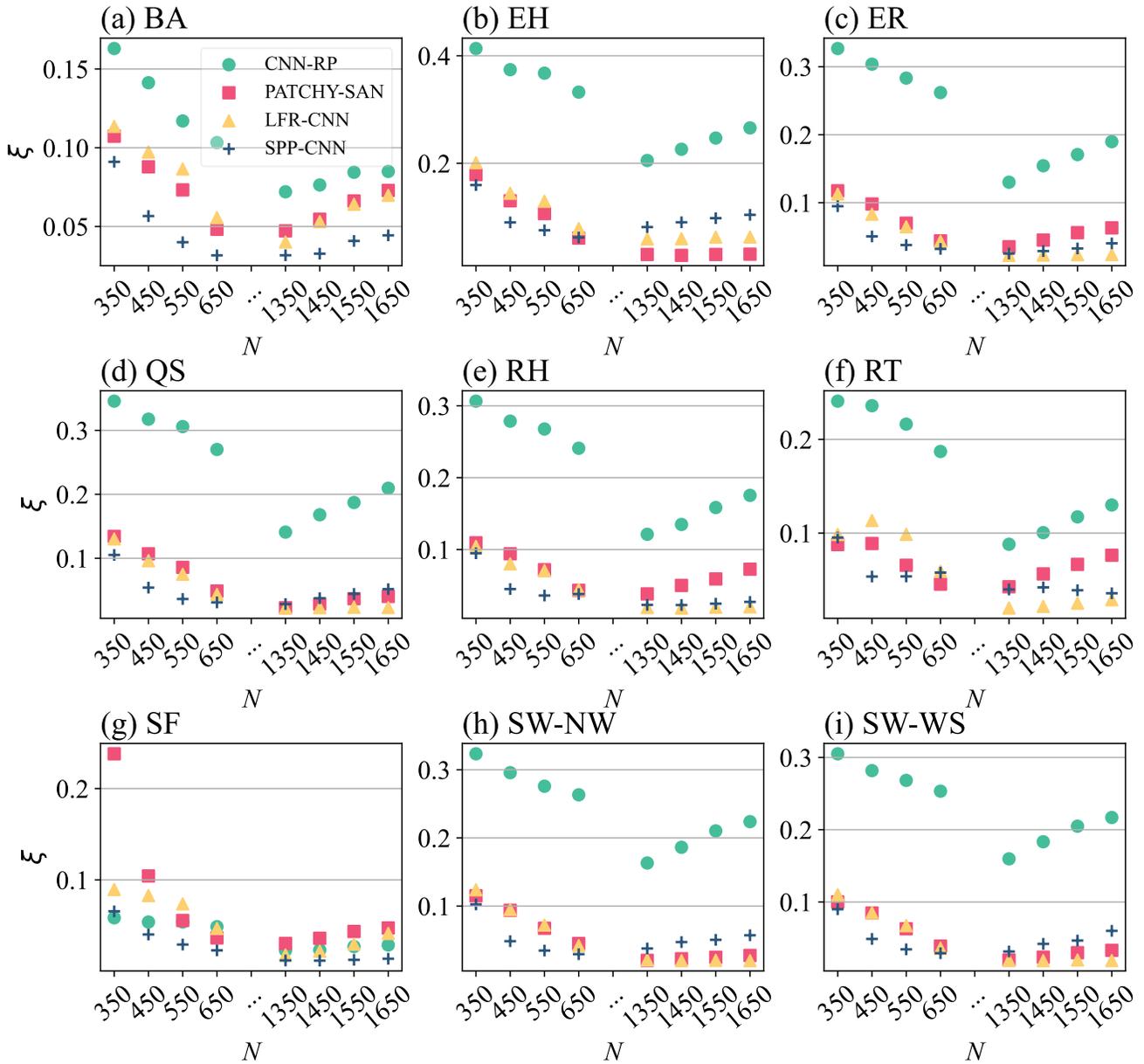

Fig. S7: Prediction errors obtained by SPP-CNN, CNN-RP, PATCHY-SAN, and LFR-CNN for unseen network sizes (UNS). Connectivity robustness of undirected networks under maximum-degree node attacks is predicted.

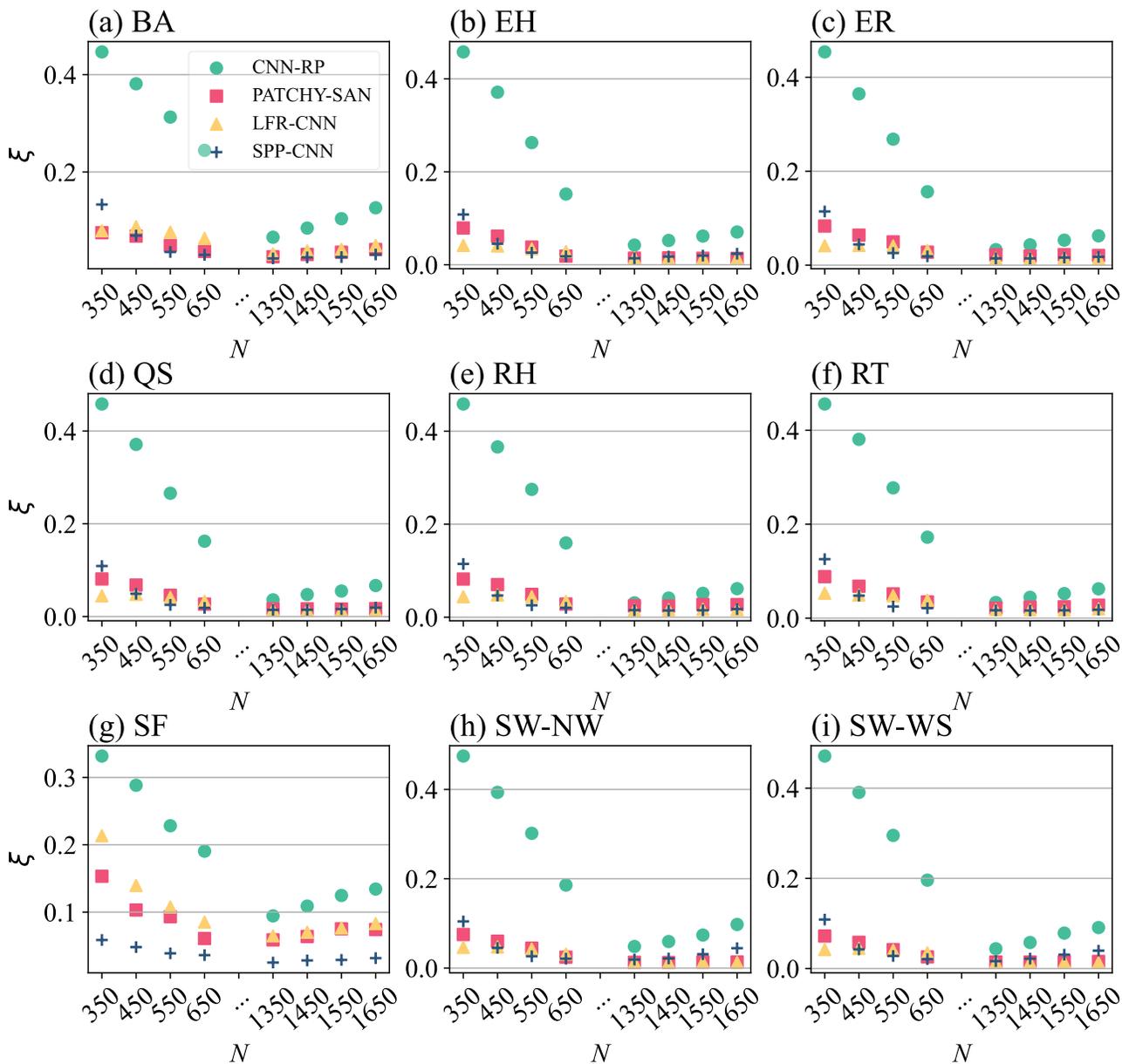

Fig. S8: Prediction errors obtained by SPP-CNN, CNN-RP, PATCHY-SAN, and LFR-CNN for unseen network sizes (UNS). Controllability robustness of undirected networks under random node attacks is predicted.

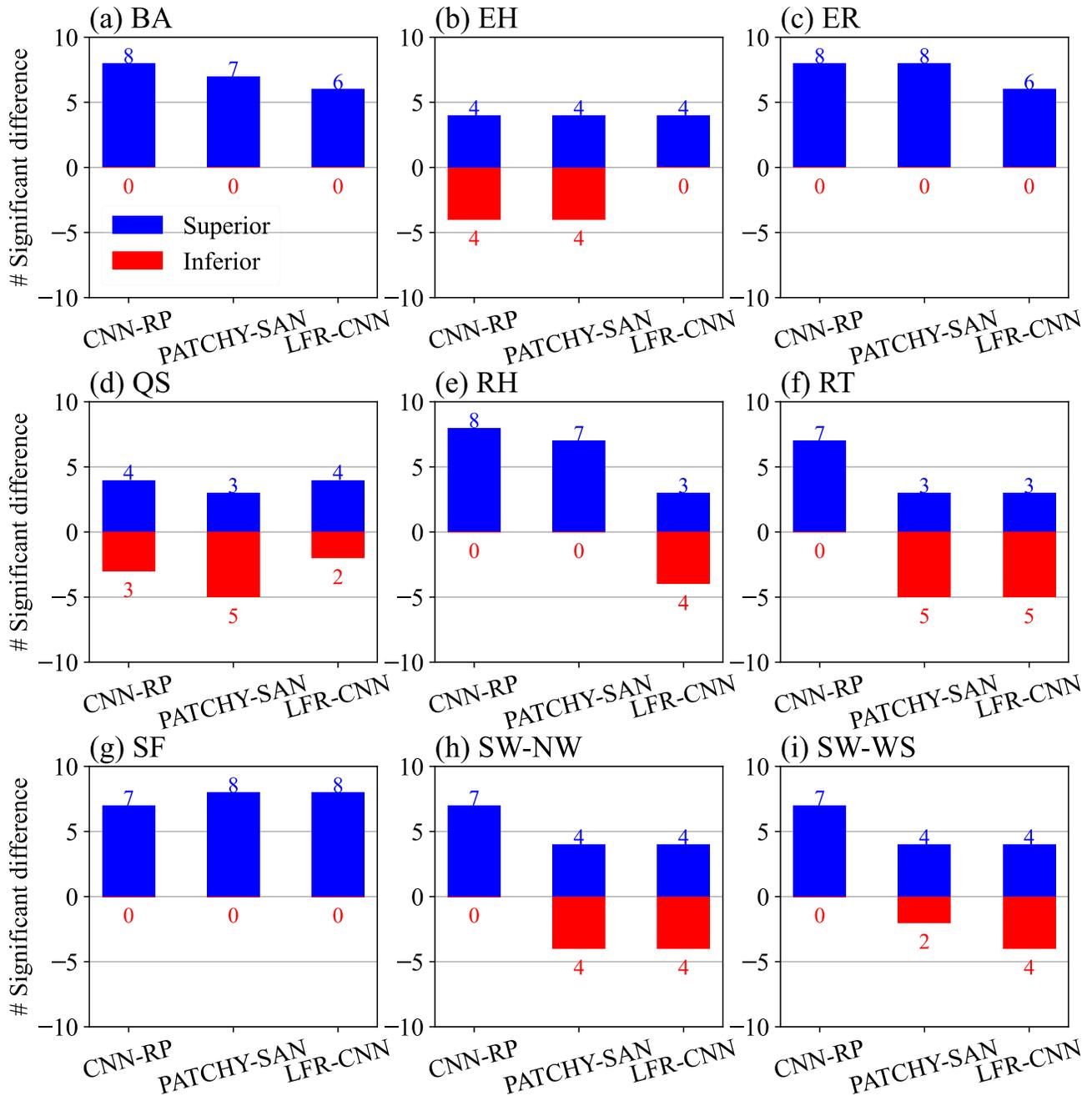

Fig. S9: Numbers of superiors and inferiors obtained by SPP-CNN, compared to each one of CNN-RP, PATCHY-SAN, and LFR-CNN. Connectivity robustness of directed networks under maximum-degree node attacks is predicted.

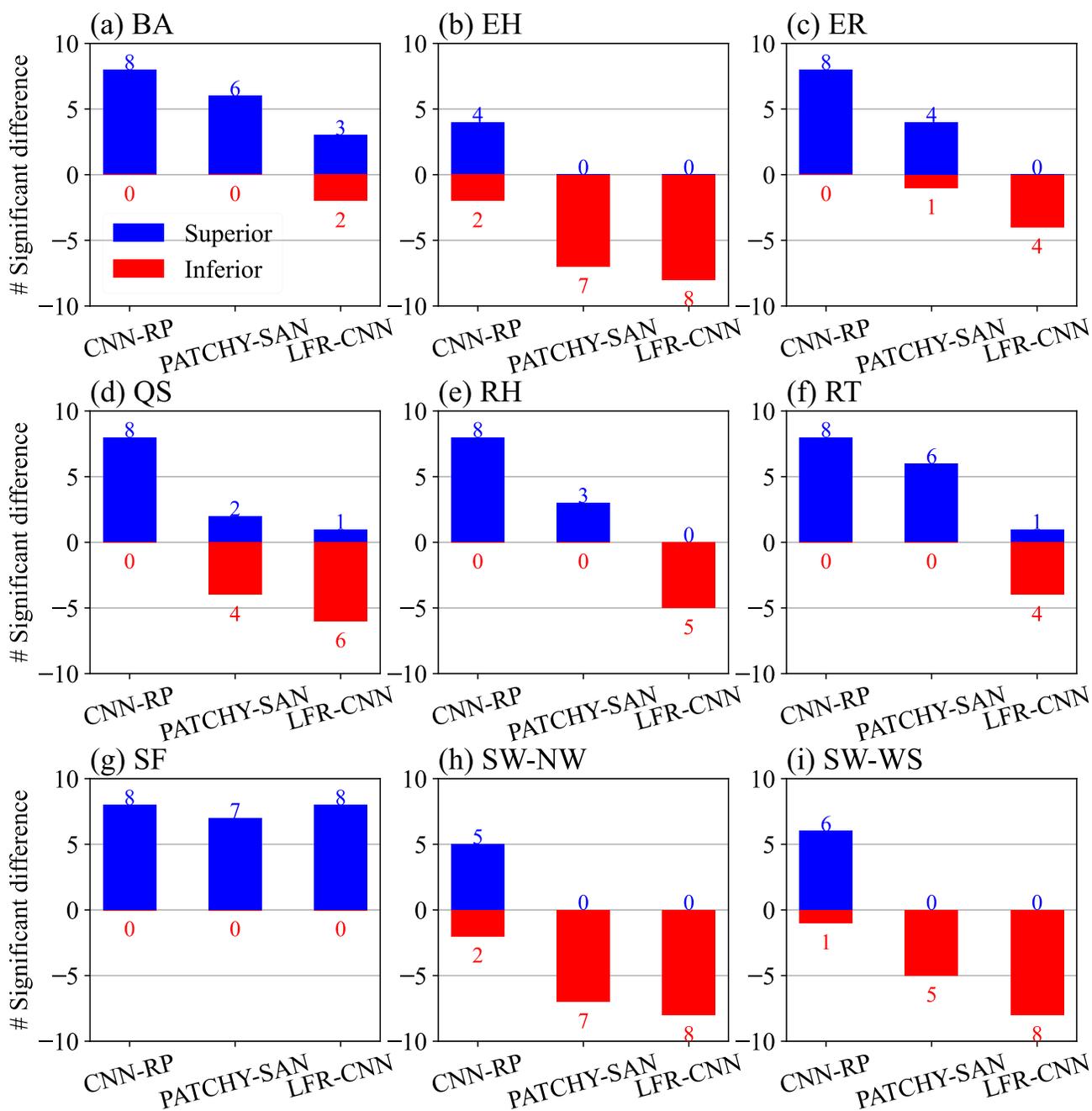

Fig. S10: Numbers of superiors and inferiors obtained by SPP-CNN, compared to each one of CNN-RP, PATCHY-SAN, and LFR-CNN. Controllability robustness of directed networks under random node attacks is predicted.

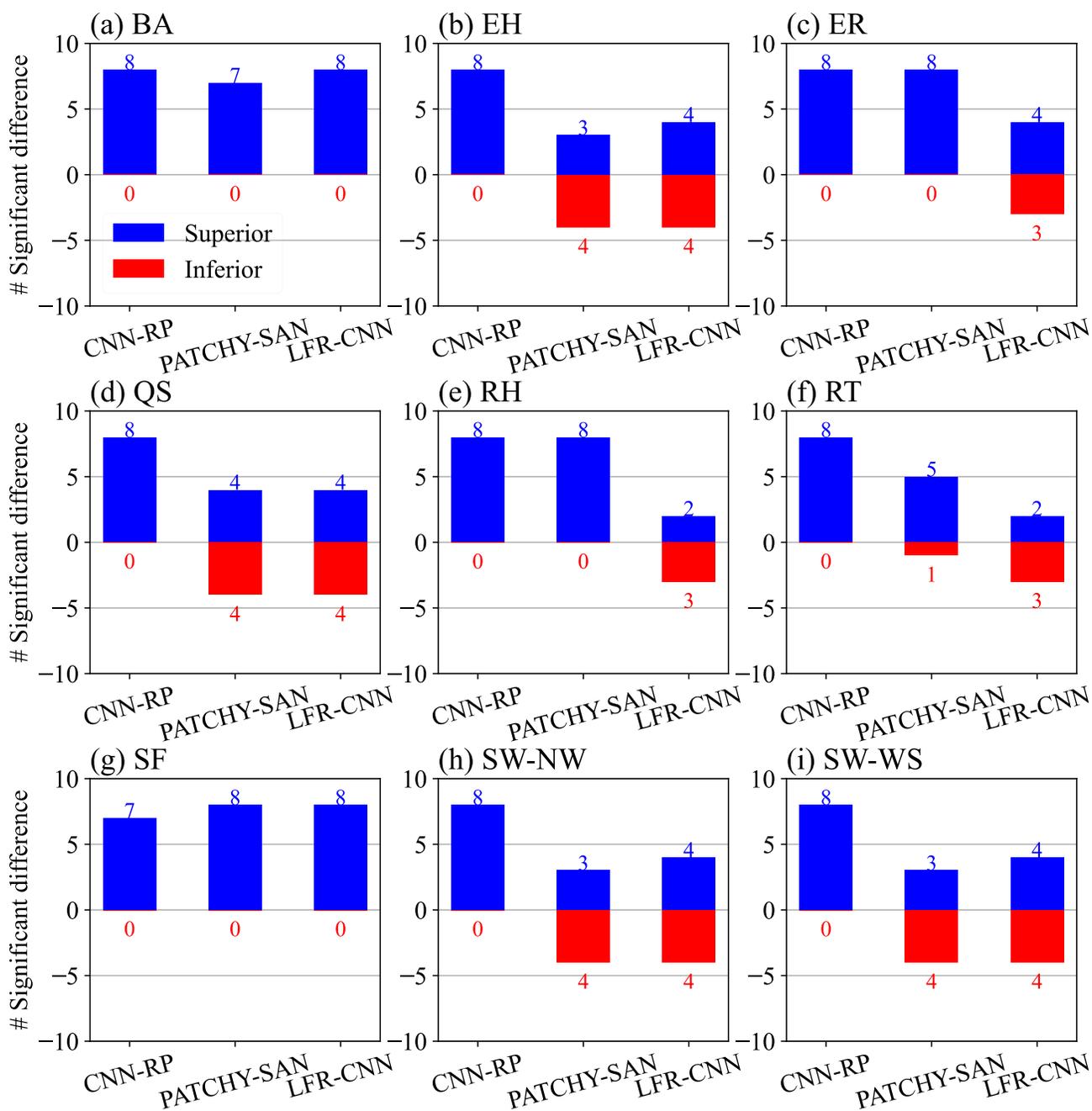

Fig. S11: Numbers of superiors and inferiors obtained by SPP-CNN, compared to each one of CNN-RP, PATCHY-SAN, and LFR-CNN. Connectivity robustness of undirected networks under maximum-degree node attacks is predicted.

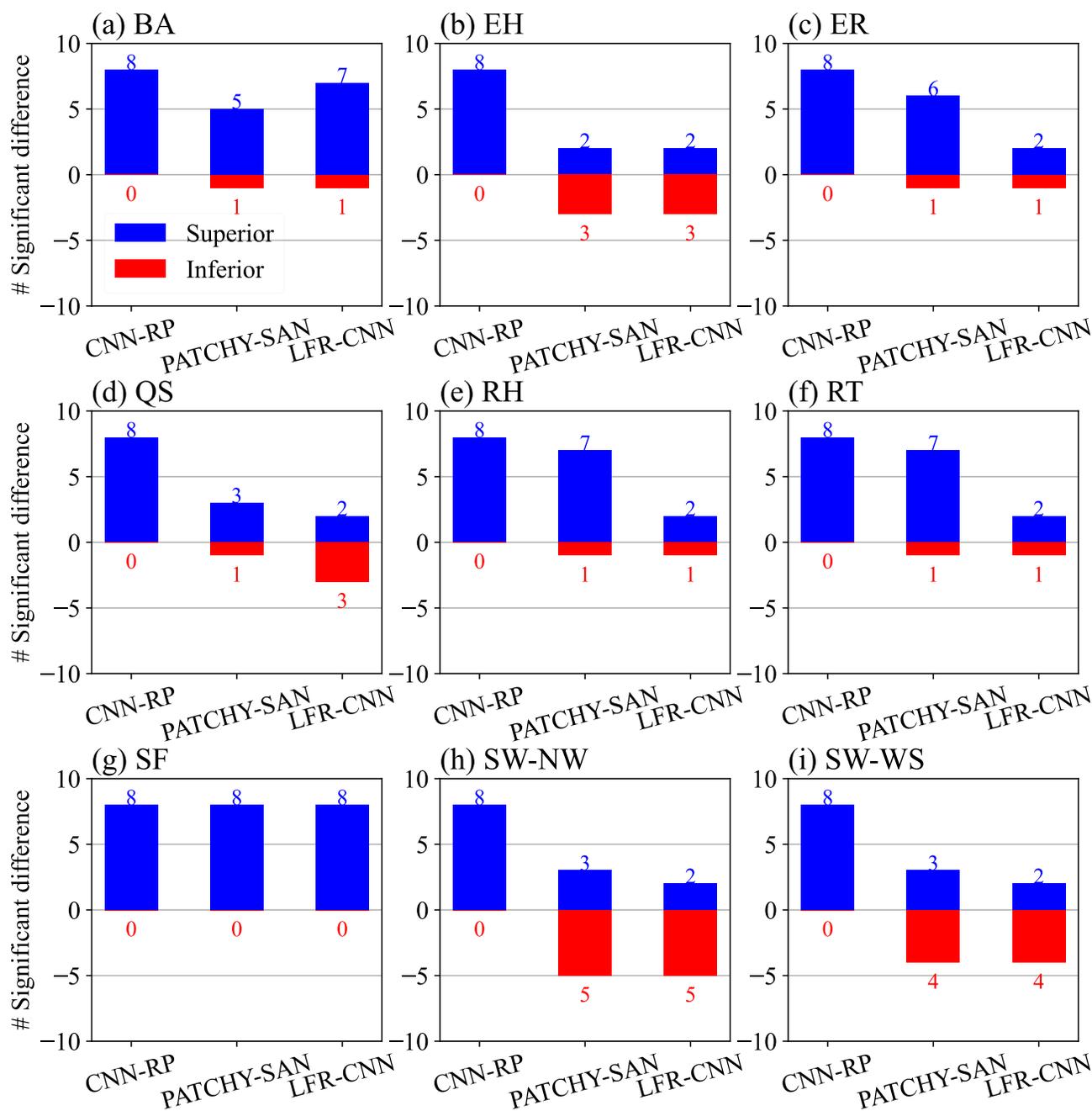

Fig. S12: Numbers of superiors and inferiors obtained by SPP-CNN, compared to each one of CNN-RP, PATCHY-SAN, and LFR-CNN. Controllability robustness of undirected networks under random node attacks is predicted.

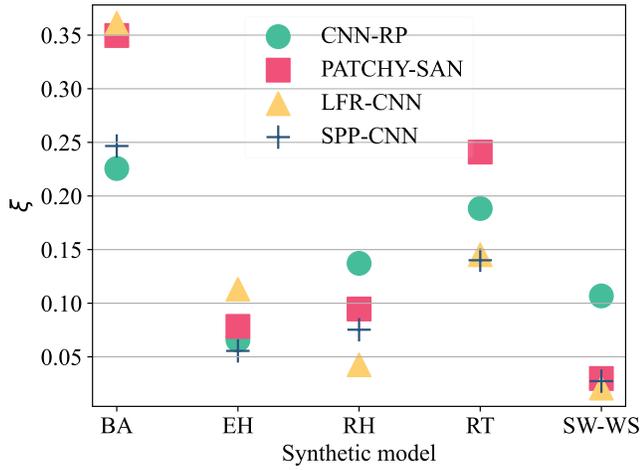

(a) Connectivity robustness of directed networks under maximum-degree node attacks.

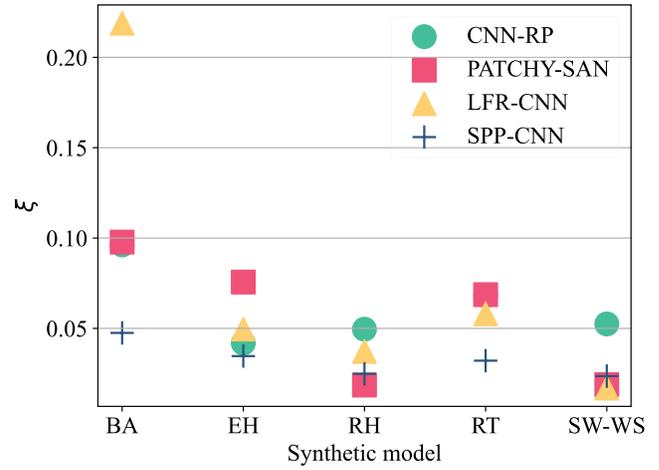

(b) Connectivity robustness of directed networks under random node attacks.

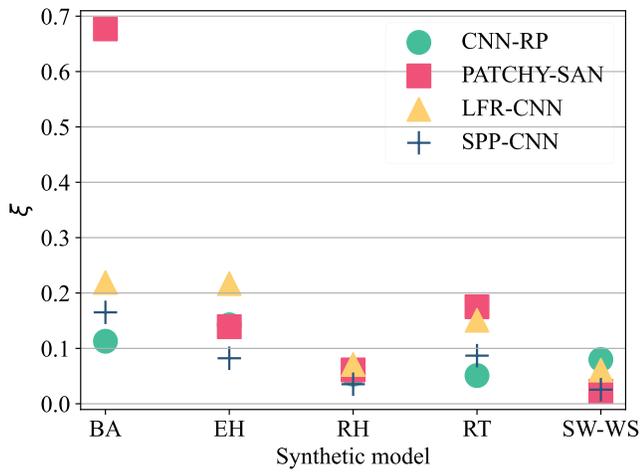

(c) Connectivity robustness of undirected networks under maximum-degree node attacks.

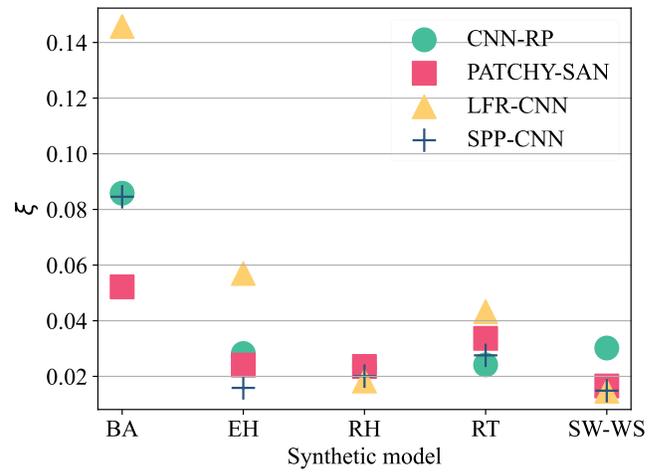

(d) Controllability robustness of undirected networks under random node attacks.

Fig. S13: Prediction errors obtained by SPP-CNN, CNN-RP, PATCHY-SAN, and LFR-CNN for unseen network topology (UNT).

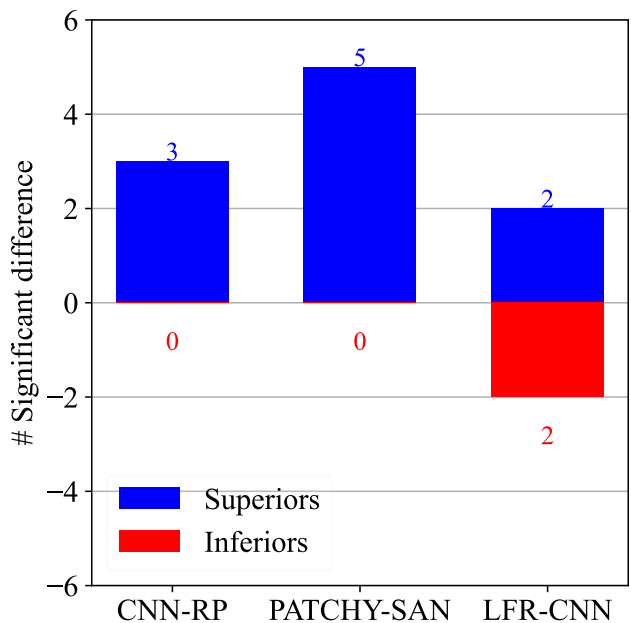

(a) Connectivity robustness of directed networks under maximum-degree node attacks.

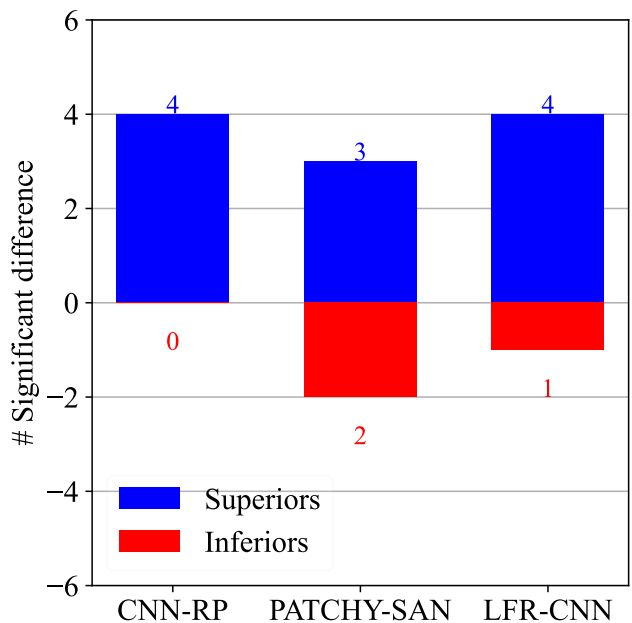

(b) Controllability robustness of directed networks under random node attacks.

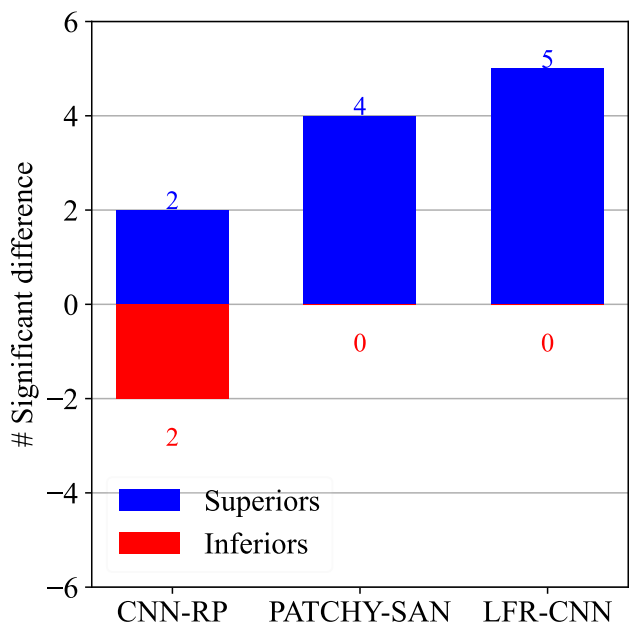

(c) Connectivity robustness of undirected networks under maximum-degree node attacks.

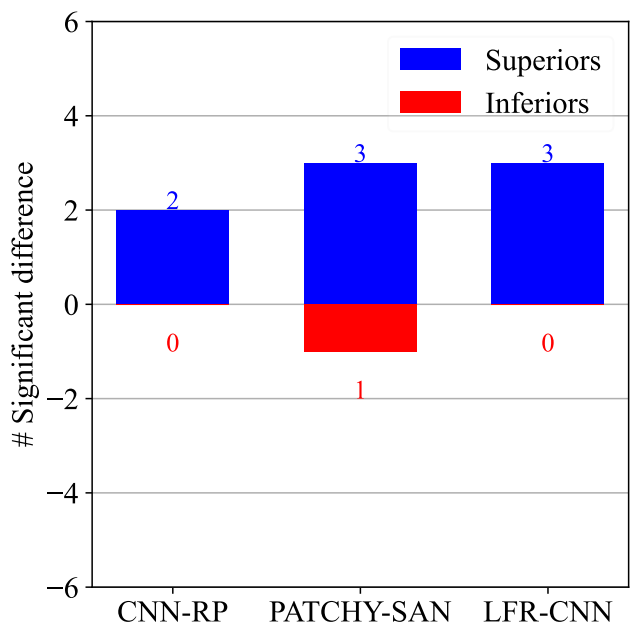

(d) Controllability robustness of undirected networks under random node attacks.

Fig. S14: Numbers of superiors and inferiors obtained by SPP-CNN, compared to each one of CNN-RP, PATCHY-SAN, and LFR-CNN for unseen network topology (UNT).



\end{document}